\PassOptionsToPackage{dvipsnames,table}{xcolor}
\documentclass{openwam}

\usepackage[utf8]{inputenc}
\usepackage{enumitem}
\usepackage{url}
\usepackage{amsmath}
\usepackage{amsfonts}
\usepackage{dsfont} 
\usepackage{nicefrac}
\usepackage{tablefootnote}
\usepackage{tabularx}
\usepackage{algorithm}
\usepackage{algpseudocode}
\usepackage{fontawesome5}
\newsavebox{\owmgradienttext}
\newsavebox{\owmgradientform}
\newdimen\owmgradientpad
\DeclareRobustCommand{\onlinewmgradient}[1]{%
  \leavevmode\begingroup
  \sbox{\owmgradienttext}{#1}%
  \pgfmathsetmacro{\owmw}{\wd\owmgradienttext/1bp}%
  \pgfmathsetmacro{\owmh}{(\ht\owmgradienttext+\dp\owmgradienttext)/1bp}%
  \pgfmathsetmacro{\owmd}{\dp\owmgradienttext/1bp}%
  \pgfmathsetmacro{\owml}{\owmw*sin(112)+\owmh*abs(cos(112))}%
  \pgfmathsetmacro{\owmxa}{\owmw/2-\owml*sin(112)/2}%
  \pgfmathsetmacro{\owmya}{\owmh/2-\owmd-\owml*cos(112)/2}%
  \pgfmathsetmacro{\owmxb}{\owmw/2+\owml*sin(112)/2}%
  \pgfmathsetmacro{\owmyb}{\owmh/2-\owmd+\owml*cos(112)/2}%
  \immediate\pdfobj{<< /ShadingType 2 /ColorSpace /DeviceRGB
    /Coords [\owmxa\space\owmya\space\owmxb\space\owmyb]
    /Function << /FunctionType 3 /Domain [0 1]
      /Functions [
        << /FunctionType 2 /Domain [0 1]
           /C0 [0.13333333 0.47843137 0.51372549]
           /C1 [0.23529412 0.40784314 0.70196078] /N 1 >>
        << /FunctionType 2 /Domain [0 1]
           /C0 [0.23529412 0.40784314 0.70196078]
           /C1 [0.47450980 0.38039216 0.67843137] /N 1 >>]
      /Bounds [0.48] /Encode [0 1 0 1] >> /Extend [true true] >>}%
  \edef\owmshading{\the\pdflastobj}%
  \owmgradientpad=.2em\relax
  \setbox\owmgradientform=\hbox{%
    \kern\owmgradientpad
    \pdfliteral{q 7 Tr}%
    \copy\owmgradienttext
    \kern-\wd\owmgradienttext
    \pdfliteral{/OWMGradient sh Q}%
    \kern\wd\owmgradienttext
    \kern\owmgradientpad
  }%
  \ht\owmgradientform=\dimexpr\ht\owmgradienttext+\owmgradientpad\relax%
  \dp\owmgradientform=\dimexpr\dp\owmgradienttext+\owmgradientpad\relax%
  \immediate\pdfxform resources{/Shading << /OWMGradient \owmshading\space 0 R >>}\owmgradientform
  \setbox\owmgradientform=\hbox{\kern-\owmgradientpad\pdfrefxform\pdflastxform\kern-\owmgradientpad}%
  \wd\owmgradientform=\wd\owmgradienttext
  \ht\owmgradientform=\ht\owmgradienttext
  \dp\owmgradientform=\dp\owmgradienttext
  \box\owmgradientform
  \endgroup
}

\DeclareRobustCommand{\onlinewmtitleicon}{%
  \begingroup
  \pdfimageresolution=72\relax
  \raisebox{-0.10em}{\includegraphics[
    height=0.79em,trim={144bp 212bp 141bp 202bp},clip
  ]{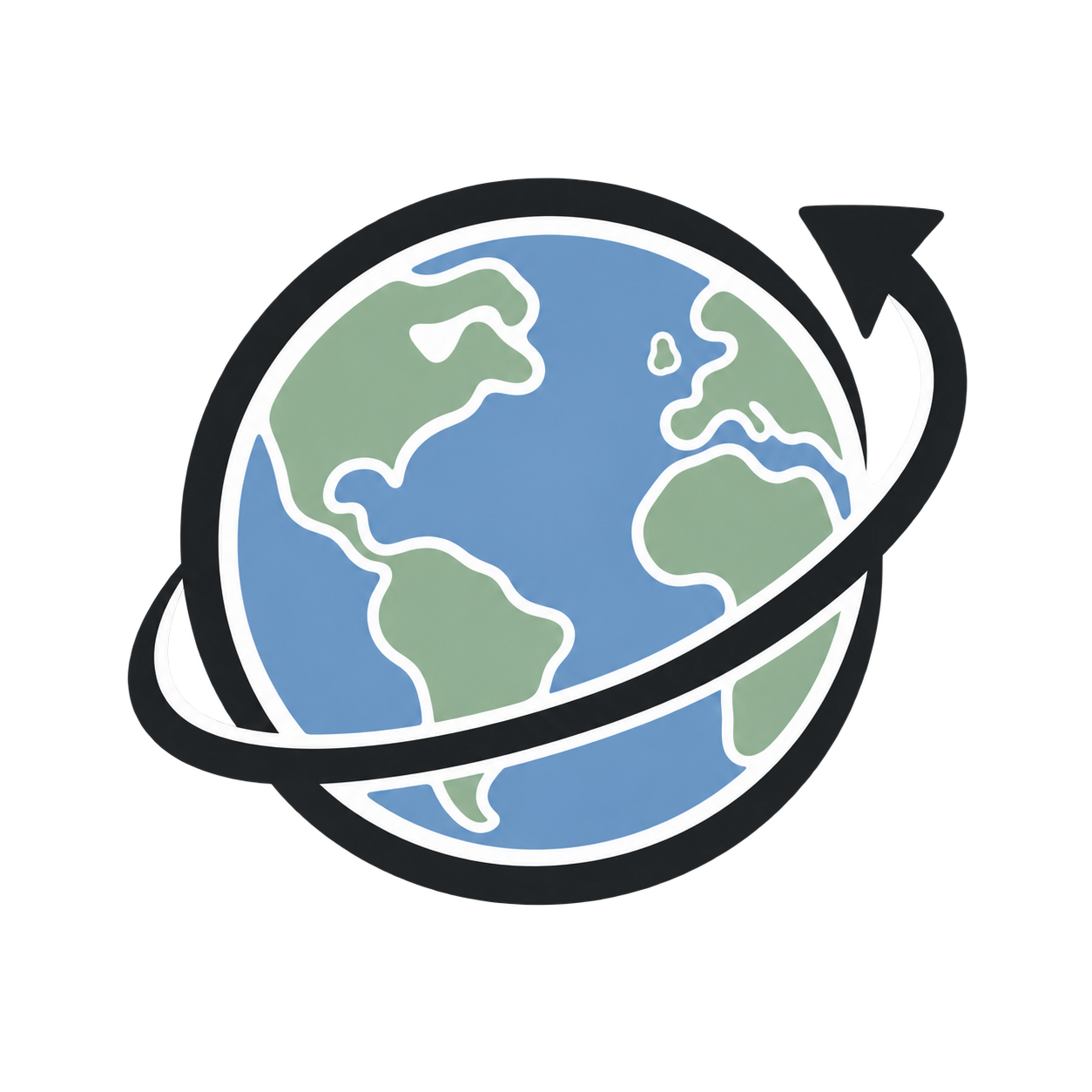}}%
  \endgroup\nobreak\hspace{0.3em}}
\colorlet{onlinewmcite}{MidnightBlue}
\definecolor{onlinewmrefred}{HTML}{FF0000}
\definecolor{onlinewmprojectpink}{HTML}{D63384}
\definecolor{wamtitle}{HTML}{000000}
\definecolor{wambg}{HTML}{EAF2FF}
\hypersetup{
  colorlinks=true,
  pdfborder={0 0 0},
  citecolor=onlinewmcite,
  linkcolor=onlinewmrefred,
  urlcolor=onlinewmcite,
  pdftitle={OnlineWM: Causality-Aware Active Online Learning for Effective World Modeling},
  pdfauthor={Yikun Miao, Fangqi Zhu, Quanxin Shou, Xiaoyi Pang, Zhengyang Yan, Junhao Li, Haodong Wang, Zicong Hong, Song Guo}
}
\setcitestyle{authoryear,round,semicolon,aysep={,},yysep={;}}

\DeclareRobustCommand{\onlinewmmetadataicon}[1]{%
  \textcolor{black}{\makebox[1.25em][c]{#1}}}
\newcommand{\homepage}[1]{%
  \metadata[\onlinewmmetadataicon{\faGlobe}~Project Page]{#1}}

\renewcommand{\correspondence}[1]{%
  \contribution[\dagger]{\onlinewmmetadataicon{\faIcon[regular]{envelope}}~%
    {\sffamily\bfseries Correspondence:} #1}}

\crefname{figure}{\textcolor{black}{Fig.}}{\textcolor{black}{Figs.}}
\Crefname{figure}{\textcolor{black}{Fig.}}{\textcolor{black}{Figs.}}
\crefname{table}{\textcolor{black}{Tab.}}{\textcolor{black}{Tabs.}}
\Crefname{table}{\textcolor{black}{Tab.}}{\textcolor{black}{Tabs.}}
\crefname{section}{\textcolor{black}{Sec.}}{\textcolor{black}{Secs.}}
\Crefname{section}{\textcolor{black}{Sec.}}{\textcolor{black}{Secs.}}
\crefname{subsection}{\textcolor{black}{Sec.}}{\textcolor{black}{Secs.}}
\Crefname{subsection}{\textcolor{black}{Sec.}}{\textcolor{black}{Secs.}}
\crefname{subsubsection}{\textcolor{black}{Sec.}}{\textcolor{black}{Secs.}}
\Crefname{subsubsection}{\textcolor{black}{Sec.}}{\textcolor{black}{Secs.}}

\DeclareRobustCommand{\onlinewmrefnumber}[1]{\begingroup
  \hypersetup{linkcolor=onlinewmrefred}\ref{#1}\endgroup}
\DeclareRobustCommand{\figref}[1]{\textcolor{black}{Fig.}~\onlinewmrefnumber{#1}}
\DeclareRobustCommand{\tabref}[1]{\textcolor{black}{Tab.}~\onlinewmrefnumber{#1}}
\DeclareRobustCommand{\secref}[1]{\textcolor{black}{Sec.}~\onlinewmrefnumber{#1}}
\DeclareRobustCommand{\figrefnum}[1]{\onlinewmrefnumber{#1}}

\newcommand{\onlinewmheadingfont}{\fontfamily{papersans}\fontseries{b}\selectfont}
\titleformat*{\section}{\Large\onlinewmheadingfont}
\titleformat*{\subsection}{\large\onlinewmheadingfont}
\titleformat*{\subsubsection}{\normalsize\onlinewmheadingfont}
\titleformat*{\paragraph}{\onlinewmheadingfont}
\renewcommand{\title}[1]{\def\titlelist{{\huge\onlinewmheadingfont\color{wamtitle}#1}}}

\patchcmd{\mymaketitle}
  {\hypersetup{linkcolor=wamtitle,citecolor=wamtitle,urlcolor=wamtitle}}
  {\hypersetup{linkcolor=onlinewmrefred,citecolor=onlinewmcite,urlcolor=onlinewmcite}}
  {}{\PackageError{onlinewm}{Could not set title-panel link colors}{Check openwam.cls.}}

\newcommand{\ie}{\textit{i.e}.}
\newcommand{\eg}{\textit{e.g}.}

\title{\onlinewmtitleicon \onlinewmgradient{OnlineWM}: Causality-Aware Active Online Learning for Effective World Modeling}
\author{Yikun~Miao}
\author{Fangqi~Zhu}
\author{Quanxin~Shou}
\author{Xiaoyi~Pang}
\author{Zhengyang~Yan}
\author{Junhao~Li}
\author{Haodong~Wang}
\author{Zicong~Hong}
\author[\dagger]{Song~Guo}
\affiliation{Department of Computer Science and Engineering\\
  The Hong Kong University of Science and Technology\\Hong Kong SAR, China}
\correspondence{\email{songguo@cse.ust.hk}}
\homepage{\href{https://OnlineWM.github.io/}{\textcolor{onlinewmprojectpink}{\textrm{https://OnlineWM.github.io}}}}

\abstract{
Generative world models aim to predict future states conditioned on actions, where action controllability is fundamental for reliable dynamics modeling. While recent efforts leverage simulator-generated data to enhance this capability, existing training pipelines face two fundamental limitations. First, static offline data collection leads to a distribution misalignment between training sets and the model’s evolving error patterns, failing to resolve critical \textit{long-tail} scenarios where dynamics predictions remain unreliable. Second, the standard objective of minimizing observational discrepancy often encourages the model to exploit spurious correlations instead of capturing the underlying action-effect causality. To address these limitations, we propose \textbf{OnlineWM}, an online training framework that continuously improves world modeling through active simulator interaction and causality-aware optimization. OnlineWM introduces two key innovations: (1) \emph{Active Online Learning}: Instead of using fixed datasets, OnlineWM adaptively queries the simulator for new interaction sequences that target the model's current predictive weaknesses, ensuring high-utility data acquisition. (2) \emph{Causality-Aware Fine-Tuning}: We propose a counterfactual learning strategy that contrasts the outcomes of different actions from identical states, forcing the model to attribute state transitions to specific actions rather than ambient environmental evolution, thereby grounding its predictions in reliable causal mechanisms. By integrating active data acquisition with causal optimization, OnlineWM establishes a closed-loop refinement process that ensures the model is both robust to diverse scenarios and precise in its causal attribution. Extensive experiments demonstrate that OnlineWM significantly enhances action controllability and generalizes effectively to unseen domains, suggesting the learning of physically-grounded causal dynamics rather than simple visual patterns.
}

\begin{document}
\maketitle

\section{Introduction}
Recent advances in video generation have enabled the synthesis of controllable, high-fidelity, and temporally coherent visual content~\citep{wan2025wan,wu2025hunyuanvideo,zheng2024open,lin2024open}, opening a promising avenue toward modeling the physical world directly in pixel space. Building upon this progress, generative world models~\citep{bruce2024genie,zhu2025irasim,yu2025gamefactory,he2025matrix,team2026advancing,ha2018world} have emerged as a promising route for forward dynamics prediction, which formulates world modeling as an action-conditioned video prediction problem: given a history of frames and an executed action (\eg, player movements or camera rotations), the model is expected to synthesize future frames that are consistent with the underlying transition dynamics. This paradigm shows broad applicability in robotics~\citep{zhu2025wmpo,shou2026halo}, autonomous driving~\citep{agarwal2025cosmos}, and game development~\citep{yu2025gamefactory,he2025matrix}.

\begin{figure}[t]
    \centering
    \includegraphics[width=0.9\linewidth]{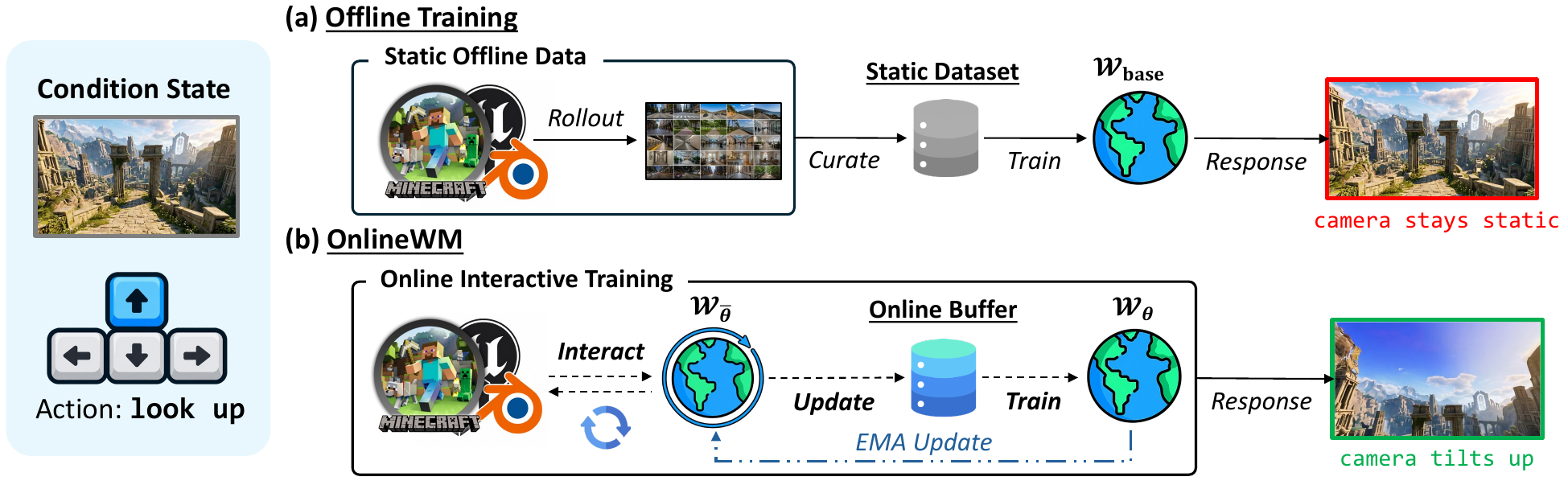}
    \caption{\textbf{Offline \textit{vs.} online training of action-controllable world models.}
    Given a condition state and an action (\texttt{look up}), a world model is tasked to predict the next observation.
    \textbf{(a) Prior work} trains $\mathcal{W}_{\text{base}}$ on a static dataset curated offline from simulators, which suffers from inherent action imbalance and leaves long-tail action--scene pairings undersampled, causing the prediction to ignore the action.
    \textbf{(b) OnlineWM (ours)} maintains an EMA copy $\mathcal{W}_{\bar\theta}$ that actively interacts with the simulator; the resulting rollouts are pushed into an online buffer that drives the training of $\mathcal{W}_\theta$, whose updates flow back to $\mathcal{W}_{\bar\theta}$ via EMA, yielding correct action grounding.}
    \label{fig:teaser}
\end{figure}

Despite their potential, state-of-the-art world models still exhibit significant inaccuracies in action controllability (\ie, the model's ability to precisely manifest the visual consequences of a specific control signal while maintaining temporal consistency), particularly when the model encounters actions that are underrepresented in its training distribution or represent out-of-distribution (OOD) pairings of actions and scene contexts. For example, as illustrated in \figref{fig:teaser}, a world model may fail to faithfully simulate a \texttt{look up} command not because the action itself is inherently difficult to model, but because it is rarely co-observed with \emph{walking trajectories} in the training distribution. Such failures are largely rooted in the inherent biases of large-scale real-world datasets used for pre-training, where action distributions are highly imbalanced and sparse, and action annotations can be noisy or inaccurate. To alleviate these data-driven biases, recent methods have increasingly relied on supervised fine-tuning on controllable simulator-generated data~\citep{team2026advancing,he2025matrix,yu2025gamefactory,sun2025worldplay}. However, this ``collect-then-train'' paradigm faces two inherent structural bottlenecks that limit its effectiveness in complex dynamics modeling.

First, existing data collection pipelines typically rely on offline datasets generated via fixed heuristics, such as uniform~\citep{yu2025gamefactory} or quality-biased sampling~\citep{he2025matrix}. While they provide broad coverage of collected trajectories, such a decoupling of data acquisition from the model’s learning state leads to a misalignment between fixed training distributions and the model’s evolving error patterns. Consequently, the training process suffers from redundant transitions while the critical long-tail scenarios—where the model’s current dynamics prediction remains unreliable—remain undersampled. Second, the reliance on standard visual prediction loss functions induces a \textit{causal ambiguity} during optimization. That is, the model may generate high-fidelity videos that nonetheless ignore the action control signals and follow common motion patterns, failing to achieve true action controllability. Ultimately, these limitations underscore a critical gap: the lack of an interactive mechanism that can adaptively align data acquisition with the model's evolving weaknesses while enforcing rigorous causal grounding.

To bridge the gap, we propose \textbf{OnlineWM}, an online training framework that transforms world model learning from static supervision to a closed-loop, active and causality-driven process. The core of OnlineWM lies in two synergistic components. First, we introduce \emph{Active Online Learning} (AOL) that proactively queries simulators for high-gain samples with balanced difficulty (\ie, novel to the current model while remaining learnable~\citep{hughes2024open}). To this end, we propose a difficulty decomposition strategy that separates the model's predictive errors into scene-level and action-level components, which allows the framework to explicitly identify and prioritize scenarios where learning is effective for both scene context modeling and transition dynamics. By maintaining a balanced contribution from both dimensions during data acquisition, OnlineWM ensures that the training process does not become biased toward pure visual complexity, and remains focused on the critical task of grounding actions within appropriate physical contexts. Second, to ensure accurate learning of causal dynamics from the crafted dataset, we develop \emph{Causality-Aware Fine-Tuning} (CFT). By constructing explicit counterfactual scenarios and optimizing for the model's ability to discriminate~\citep{miao2024hierarchical} between divergent action outcomes within identical states, CFT ensures that the latent space disentangles action-induced perturbations from common ambient scene evolution. Together, these components enable OnlineWM to achieve superior action controllability and robust generalization in complex scenarios.

To conclude, our contributions can be summarized as:
\begin{itemize}[leftmargin=1em]
    \item We propose OnlineWM, an online training framework that shifts world model training from passive supervision to an interactive, causality-driven refinement process. By establishing a dynamic feedback loop between the world model's current state and the simulator, OnlineWM actively aligns the model's internal dynamics with the causal laws of the environment.
    \item We design two synergistic components that expose and rectify the model's causal deficiencies: AOL isolates action-specific failures from environmental complexity, enabling the targeted identification of samples with high epistemic value; and CFT leverages these samples for counterfactual training, forcing the model to learn the reliable causal relationship between actions and visual changes.
    \item Extensive experiments prove that OnlineWM effectively improves state-of-the-art world models~\citep{sun2025worldplay} in both action controllability and visual quality. The results also demonstrate OnlineWM's effectiveness in general domains with long-tail or rare action-scene pairings, suggesting that it can learn generalizable knowledge about action-conditioned world dynamics.
\end{itemize}

\section{Related Work}
\paragraph{Generative World Models.}
Breakthroughs in visual generative modeling~\citep{peebles2023scalable,lipman2022flow,podell2023sdxl} have enabled the synthesis of high-quality visual content and motivated early attempts to simulate interactive environments with generative models~\citep{valevski2024diffusion,alonso2024diffusion,bruce2024genie}. Recent advances in video generation~\citep{wu2025hunyuanvideo,wan2025wan,zheng2024open,lin2024open} further improve temporal coherence, visual fidelity, and controllability, providing a strong foundation for generative world models to continuously improve generation quality~\citep{bruce2024genie,parkerholder2024genie2,genie3}. Unlike general video generation, generative world models aim to predict future observations conditioned on both past visual states and agent actions, requiring the model to capture action-conditioned transition dynamics. More recent works~\citep{yu2025gamefactory,he2025matrix,team2026advancing,mao2025yume,li2025hunyuan,tang2025hunyuan,hong2025relic} further improve generative world models by constructing or selecting datasets with accurate action annotations, balanced action distributions, and high-quality action trajectories. These efforts substantially improve the data foundation for action-conditioned generation, but the resulting models still often struggle with precise action following under rare or out-of-distribution actions, or in visually complex scenes.

\paragraph{Improving Action Controllability for Generative World Models.}
Since generated future states should faithfully reflect the executed actions to support downstream applications~\citep{zhu2025wmpo}, accurate action control is central to generative world modeling. Recent works improve action controllability either by strengthening the action representation or through dedicated post-training procedures. For instance, WorldCam~\citep{nam2026worldcam} adopts camera pose as a unifying geometric representation to align user actions with 3D camera motion, while WorldCompass~\citep{wang2026worldcompass} introduces reward-based post-training built upon the DiffusionNFT framework~\citep{zheng2025diffusionnft} to jointly improve action accuracy and visual quality. While effective, these methods treat the training data and learning process as fixed, leaving the model passive with respect to its own learning dynamics. A complementary perspective from active world model learning~\citep{kim2020active} emphasizes that accurate world modeling benefits from adapting data acquisition to the model's current learning state, directing exploration toward dynamics that are complex yet learnable. It is also relevant to generative world models: action-controllability errors can vary across scenes, actions, and training stages, making fixed offline data collection less effective for targeting the model's current deficiencies. Building on this perspective, OnlineWM combines closed-loop simulator interaction with causality-aware training to improve action-conditioned dynamics modeling.

\section{Methodology}
\begin{figure}[t]
    \centering
    \includegraphics[width=\linewidth]{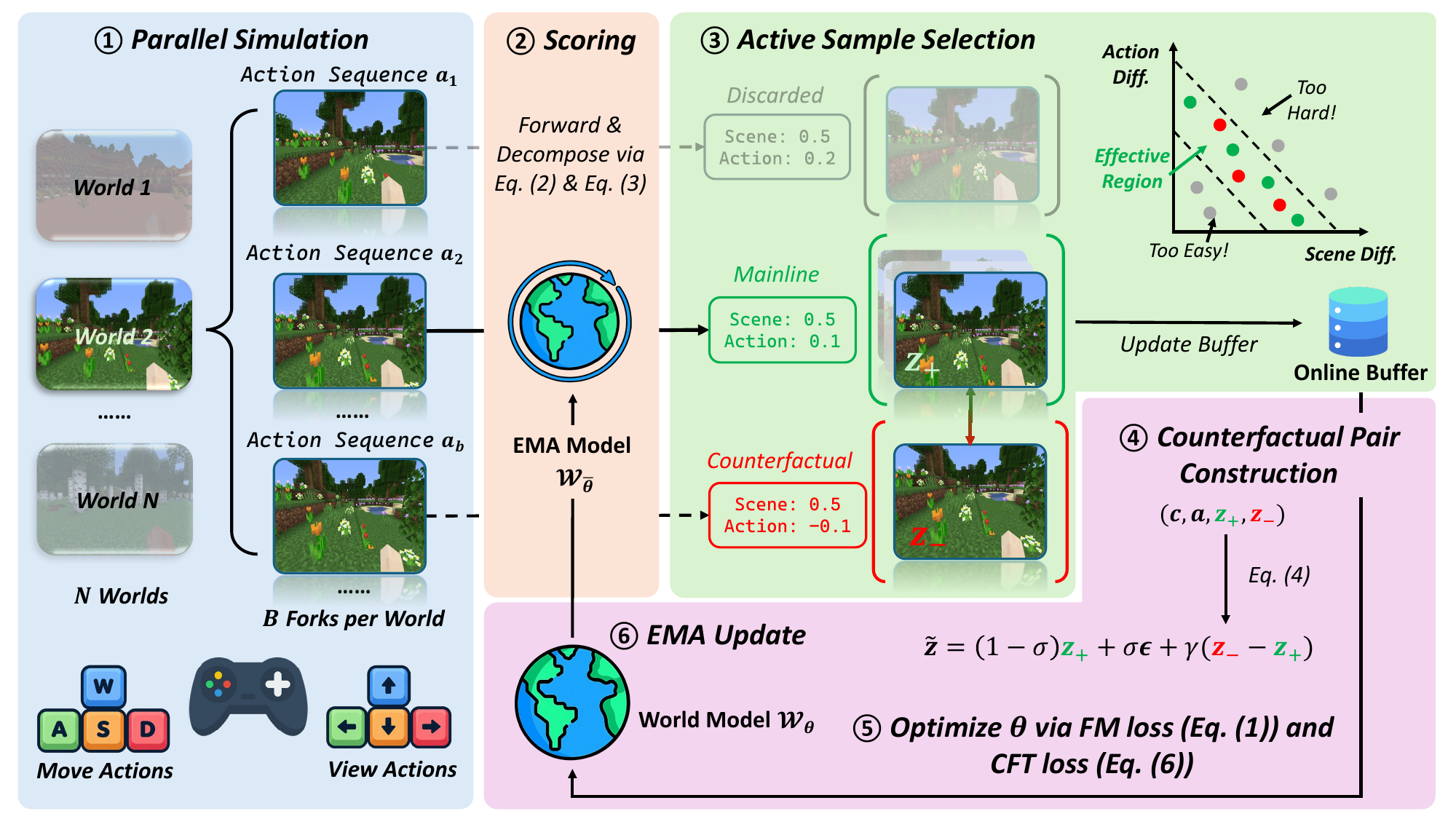}
    \caption{\textbf{OnlineWM} trains a world model with active, causality-aware online interaction with a simulator. In each collection round, all $N$ simulators generate $B$ forks from the current state. Then, the world model evaluates each fork sample and decomposes the difficulties into scene- and action-level components to select the most effective one in each world along with its most similar negative sample. The selected fork is then continued for $L_\text{main}$ steps and pushed into the buffer. In each training round, the world model draws samples from the buffer and optimizes with standard generation and causality-aware losses.}
    \label{fig:method}
\end{figure}

\subsection{Preliminaries}
\paragraph{World Modeling} 
Following representative works such as Genie~\citep{bruce2024genie}, we frame world modeling as an autoregressive video prediction problem $p(\mathbf{z}_{1:N}) = \prod_{t=1}^N p(\mathbf{z}_t | \mathbf{z}_{\leq t-1}, \mathbf{a}_t, c)$, where $\mathbf{z}_t$ and $\mathbf{a}_t$ represent the latent representation of the world state and the action taken at time $t$, while $c$ represents shared global conditioning information such as the textual prompt.

\paragraph{Autoregressive Generation with Flow-based Models}
We adopt the autoregressive Diffusion Transformer (DiT)~\citep{peebles2023scalable} architecture of HY-World 1.5~\citep{sun2025worldplay,wu2025hunyuanvideo} as our base world model, which learns to generate the latent video chunk by chunk (4 latent frames per chunk, corresponding to 16 raw frames) with chunk-wise causal attention. The model is optimized by the flow matching (FM) objective~\citep{lipman2022flow} with velocity prediction:
\begin{equation}
  \mathcal{L}_{\mathrm{FM}}(\theta) \;=\; \mathbb{E}_{t,\mathbf{z}_t,\boldsymbol{\epsilon},k,\mathbf{c}_t}\!\left[\,\big\lVert\, \mathbf{v}_\theta(\mathbf{z}_t^k, k, \mathbf{c}_t) - (\boldsymbol{\epsilon}-\mathbf{z}_t)\,\big\rVert_2^{\,2}\,\right], \quad \mathbf{z}_t^k = (1-\sigma_k)\,\mathbf{z}_t + \sigma_k\,\boldsymbol{\epsilon}.
  \label{eq:FM}
\end{equation}
In Eq.~\ref{eq:FM}, $t \in \{1,\dots,N\}$ denotes the chunk index and $\mathbf{z}_t$ denotes the clean video latent chunk encoded by the 3D VAE~\citep{kingma2013auto} at index $t$. The clean latent is corrupted by Gaussian noise $\boldsymbol{\epsilon}\!\sim\!\mathcal{N}(\mathbf{0},\mathbf{I})$, and $\sigma_k$ is drawn from a shifted logit-normal schedule~\citep{wu2025hunyuanvideo} with timestep $k$. The condition $\mathbf{c}_t = (\mathbf{z}_{\leq t-1}, \mathbf{a}_t, c)$ includes previous latent chunks $\mathbf{z}_{\leq t-1}$, current action $\mathbf{a}_t$, and the shared global conditioning $c$. Here, action is represented by both discrete tokens (movement actions \texttt{w, a, s, d}, view actions \texttt{left, right, up, down}, or any combination) and continuous camera poses. More detailed architecture information can be found in Appendix~\ref{appendix:model-architecture}.

\subsection{Active Online Learning}
Active online learning (AOL) aims to actively acquire the most effective training samples by directing the simulator to generate rollouts according to the world model's current state. We characterize such samples as both \emph{novel} and \emph{learnable}~\citep{hughes2024open}, which allows the model to continuously acquire new information about the world while avoiding the ``white noise problem''~\citep{schmidhuber2010formal}, \ie, endlessly fixating on unlearnable stimuli~\citep{kim2020active}. As mentioned, the learning value of a training sample for generative world models depends on two complementary aspects: the scene context and the action-conditioned transition, both of which should provide learnable novelty for the current model. Motivated by this observation, we decompose sample acquisition into scene- and action-level components and select samples that maintain a balanced contribution from both.

\paragraph{Building AOL Samples} As shown in \figref{fig:method}, we launch $N$ parallel interactive simulators during training. At each collection round, we randomly sample $B$ candidate action sequences of length $L_\text{fork}$ from the action space $\mathcal{A}$ to create forks in each world. For each fork in world $n$, we then run the simulator for $L_\text{fork}$ chunks with the corresponding action and obtain the ground-truth rollouts $\mathbf{z}_{n,b}$. Note that these rollouts share the same conditioning state $\mathbf{c}$ but differ in the resulting outcomes due to executed actions. We then score the difficulty of each sample using the standard FM loss at a fixed noise timestep with one forward pass. An exponential moving average (EMA) copy $\bar{\theta}$ of the world model is used for scoring to stabilize online training:
\begin{equation}
  s_{n,b} \;=\; \big\lVert\, \mathbf{v}_{\bar{\theta}}(\tilde{\mathbf{z}}_{n,b},\, k^{\star},\, \mathbf{c}_{n,b}) - (\boldsymbol{\epsilon}-\mathbf{z}_{n,b})\,\big\rVert_2^{\,2}, \quad \tilde{\mathbf{z}}_{n,b} = (1-\sigma^{\star})\,\mathbf{z}_{n,b} + \sigma^{\star}\,\boldsymbol{\epsilon}.
  \label{eq:FM-score}
\end{equation}
Here, $\sigma^{\star}$ is a fixed scoring noise level with scheduler index $k^{\star}$. The per-fork condition $\mathbf{c}_{n,b}$ extends the shared scene state $\mathbf{c}$ with the fork-specific action $\mathbf{a}_{n,b}$, and $\boldsymbol{\epsilon}\!\sim\!\mathcal{N}(\mathbf{0},\mathbf{I})$ is a single noise tensor shared across all $B$ forks within the scoring round, so that the resulting ranking is driven purely by content differences. The score variability arises from differences in average scores across scenes and action-dependent variation within each scene. By the law of total variance, the score variability decomposes as $\mathrm{Var}_{n,b}[s_{n,b}] = \mathrm{Var}_{n}[\mathbb{E}_{b}[s_{n,b}\mid n]] + \mathbb{E}_{n}[\mathrm{Var}_{b}[s_{n,b}\mid n]]$, corresponding to between-scene variation and within-scene action variation, respectively. This motivates us to decompose $s_{n,b}$ into a scene-level mean $d_n$ and an action-level residual $\delta_{n,b}$:
\begin{align}
  d_n = \operatorname{mean}\!\left(\left\{s_{n,b}\right\}_{b=1}^{B}\right), \quad \delta_{n,b} = s_{n,b} - d_n.
\end{align}
Within each world, we select the fork \emph{whose action-level difficulty rank is inversely aligned with the world's scene-level difficulty rank}, pairing visually demanding scenes with relatively easier actions and visually simpler scenes with harder actions. This complementary pairing balances the contributions of the two difficulty dimensions in each acquired sample, preventing the training signal from being dominated by pure visual complexity while keeping every selected transition novel yet learnable for the model's current state. Starting from the selected fork, we roll out an additional $L_\text{main}$ chunks in the simulator to obtain a full training trajectory from this high-utility branch.

\subsection{Causality-Aware Fine-Tuning}
Building on the trajectories acquired by AOL, we further construct counterfactual branches for causality-aware fine-tuning (CFT), which aims to force the model to attribute observed state transitions to the executed action rather than to common ambient scene evolution. For the selected branch $b_+$ in each world, we choose the most visually similar branch as the counterfactual branch $b_-$, yielding rollouts $\mathbf{z}_+ \leftarrow \mathbf{z}_{n,b_+}$ and $\mathbf{z}_- \leftarrow \mathbf{z}_{n,b_-}$ that share an identical conditioning state $\mathbf{c}$.

\paragraph{Learning Objectives}
Following WorldCompass~\citep{wang2026worldcompass}, we adopt DiffusionNFT~\citep{zheng2025diffusionnft}, a \emph{forward-process} post-training objective to optimize the world model with counterfactual supervision. Unlike reward-based optimization, our formulation operates on explicit positive--negative sample pairs, where the positive $\mathbf{z}_+$ denotes the ground-truth continuation under the executed action and the negative $\mathbf{z}_-$ denotes a counterfactual alternative from the same conditioning state $\mathbf{c}$. To enhance causality awareness during optimization, we perturb the forward-noised latent state by injecting the negative sample with a noise-level-adaptive weight:
\begin{equation}
  \tilde{\mathbf{z}}^{k} = (1-\sigma_k)\,\mathbf{z}_+ + \sigma_k\,\boldsymbol{\epsilon} + \gamma_k\,(\mathbf{z}_- - \mathbf{z}_+),
  \label{eq:CFT-mix}
\end{equation}
where $\gamma_k = \lambda\,h\,\sigma_k(1-\sigma_k)$ controls the weight that tilts the noise toward $\mathbf{z}_-$ while keeping the boundary unchanged at $\sigma_k\!\in\!\{0,1\}$, and $h$ is a similarity metric (\eg, cosine similarity, SSIM) to modulate sample hardness. From this counterfactually perturbed state, the model is then trained to recover the positive trajectory $\mathbf{z}_+$ and repel $\mathbf{z}_-$, forcing it to attribute the predicted dynamics to the executed action rather than to the visual cues shared between $\mathbf{z}_+$ and $\mathbf{z}_-$.

Following DiffusionNFT, we denote $\mathbf{v}_\theta$ and $\mathbf{v}_{\mathrm{old}}$ as the current and EMA model velocity predictions (conditioned on $\mathbf{c}$ and timestep $k$), and define implicit positive and negative policies along with their predictions for causality-aware fine-tuning:
\begin{equation}
  \left\{
  \begin{aligned}
    \mathbf{v}^{+} &= (1-\beta)\,\mathbf{v}_{\mathrm{old}} + \beta\,\mathbf{v}_\theta, \quad & \hat{\mathbf{z}}_+ &= \tilde{\mathbf{z}}^{k} - \sigma_k\,\mathbf{v}^{+}, \\
    \mathbf{v}^{-} &= (1+\beta)\,\mathbf{v}_{\mathrm{old}} - \beta\,\mathbf{v}_\theta, \quad & \hat{\mathbf{z}}_- &= \tilde{\mathbf{z}}^{k} - \sigma_k\,\mathbf{v}^{-}.
  \end{aligned}
  \right.
  \label{eq:CFT-pred}
\end{equation}
The final CFT loss combines the two reconstructions with a similarity-based weight that allocates more supervision to similar negatives and avoids learning from samples with large discrepancies:
\begin{equation}
  \mathcal{L}_{\mathrm{CFT}}(\theta) = (1-\alpha)\,\big\lVert\hat{\mathbf{z}}_+ - \mathbf{z}_+\big\rVert_{w}^{2} + \alpha\,\big\lVert\hat{\mathbf{z}}_- - \mathbf{z}_-\big\rVert_{w}^{2},\,\alpha = h/2.
  \label{eq:CFT-loss}
\end{equation}
To stabilize the training process across noise levels and pair difficulties, each reconstruction term adopts adaptive normalization with the stop-gradient operation~\citep{wang2026worldcompass,zheng2025diffusionnft}:
\begin{equation}
  \big\lVert\hat{\mathbf{z}} - \mathbf{z}\big\rVert_{w}^{2} \;=\; \mathrm{mean}\!\left(\frac{(\hat{\mathbf{z}} - \mathbf{z})^{2}}{\mathrm{sg}\!\left[\,\mathrm{mean}\big(|\hat{\mathbf{z}} - \mathbf{z}|\big)\,\right] + \varepsilon}\right).
  \label{eq:CFT-weight}
\end{equation}

\paragraph{Remark}
Ignoring scalar factors from the reconstruction norm, the gradient of $\mathcal{L}_{\mathrm{CFT}}$ with respect to the current velocity can be written as
\begin{equation}
  \nabla_{\mathbf{v}_\theta}\mathcal{L}_{\mathrm{CFT}}
  \;\propto\;
  -\sigma_k\beta\Big[(1-\alpha)(\hat{\mathbf{z}}_+-\mathbf{z}_+)-\alpha(\hat{\mathbf{z}}_- - \mathbf{z}_-)\Big].
  \label{eq:CFT-grad}
\end{equation}
The two branches provide complementary learning signals. The positive branch pulls the predicted velocity toward reconstructing $\mathbf{z}_+$ from the augmented noisy state $\tilde{\mathbf{z}}^{k}$, which is tilted toward the counterfactual outcome. In contrast, the negative branch provides a repulsive signal that pushes the velocity away from $\mathbf{z}_-$. Consequently, CFT converts counterfactual rollouts into causal learning signals, forcing the model to distinguish the causally correct outcome from its counterfactual alternative.

\subsection{Overall Learning Objective}
\paragraph{Training Buffer} To improve sample efficiency for online training, we maintain a fixed-capacity replay buffer $\mathcal{D}$ of size $S_{\max}$ and only run the collection round every $N_{\text{collect}}$ iterations. Each sample is assigned a priority $p$ initialized as its FM score $s_{n,b}$ at collection time, together with a usage counter $n_{\text{train}}$ that increments after each draw. When $\mathcal{D}$ exceeds its capacity, we evict the sample with the smallest effective priority $\tilde{p} = p\cdot\max(0.01,\, 1 - n_{\text{train}}/N_{\max})$ to retain high-score but under-trained samples. Training samples are drawn \emph{uniformly} from $\mathcal{D}$.

For each sample drawn from the training buffer, we calculate the total loss by the weighted combination of FM (Eq.~\ref{eq:FM}) and CFT (Eq.~\ref{eq:CFT-loss}) losses and perform joint optimization:
\begin{equation}
  \mathcal{L}_{\text{total}} = \mathcal{L}_{\text{FM}} + \lambda_{\text{CFT}} \cdot \mathcal{L}_{\text{CFT}}.
\end{equation}

\section{Experiments}
\subsection{Experimental Setup}
\label{sec:experiment_setup}
\paragraph{Overall Setting}
We use HY-World~1.5~\citep{sun2025worldplay,wu2025hunyuanvideo} (with $8$B parameters) as our base world model. We initialize it from the WorldCompass~\citep{wang2026worldcompass} RL post-training checkpoint. The simulator backend uses MineStudio~\citep{cai2024minestudio} built on MineRL~\citep{guss2019minerl}, which offers flexible Minecraft environment control and distributed interaction based on Ray~\citep{moritz2018ray}. We use $4$ compute nodes, each equipped with $8$ GPUs, running $N=32$ parallel Minecraft worlds to generate rollouts of resolution $832 \times 480$ at $20$ FPS with per-frame GT camera poses. The valid action space $\mathcal{A}$ contains $9$ keyboard movements (\texttt{idle}, $\texttt{w}$, $\texttt{a}$, $\texttt{s}$, $\texttt{d}$, and the four diagonals), $5$ camera primitives (\texttt{idle}, $\texttt{left}$, $\texttt{right}$, $\texttt{up}$, $\texttt{down}$) and their combinations, resulting in $45$ discrete compound action tokens per tick. Each sampled action is held fixed for a short duration to prevent overly abrupt visual changes (see Appendix~\ref{appendix:collection} for details).

\paragraph{Training Parameters}
We train the world model for $3{,}000$ steps with the Muon optimizer~\citep{liu2025muon}. The peak learning rate is set to $2\!\times\!10^{-5}$ with a \texttt{one-cycle} schedule. We use sequence parallelism (SP) with $8$ SP groups and $2$ gradient accumulation steps, equivalent to an effective batch size of $16$ per iteration. In each AOL collection round, we fork $B=8$ candidate branches with random override probability $p_{\text{rand}}=0.2$ to prevent overfitting fixed patterns. The selected fork of length $L_{\text{fork}}=1$ is unrolled for another $L_{\text{main}}=8$ chunks. The maximum size of the replay buffer is set to $S_{\max}=96$ with $N_{\max}=N_{\text{collect}}=6$. Both the scoring and old-policy networks for CFT are EMA copies with decay factor $0.99$. We set the scoring noise level to $\sigma^{\star}=0.5$ and $\lambda=0.1, \beta=1, \lambda_{\text{CFT}}=0.2$ for the CFT loss. Full training details are listed in Appendix~\ref{appendix:training}.

\paragraph{Evaluation Metrics}
We evaluate all methods on a held-out Minecraft set, consisting of manually curated challenging scenes of 61 frames each, paired with action trajectories in which actions change more frequently than in training. We evaluate the world model from two perspectives: (i) \emph{visual quality} and (ii) \emph{action controllability}. For \emph{visual quality}, we use both VBench~\citep{huang2024vbench} metrics and conventional visual metrics, including SSIM and LPIPS~\citep{zhang2018unreasonable}, to comprehensively evaluate generation quality. For \emph{action controllability}, we first use Depth Anything~V3~\citep{lin2025depth} with Sim(3) Umeyama alignment~\citep{umeyama2002least} to estimate normalized camera poses, and then compute discrete-token control accuracy and relative pose error (RPE) against the ground truth following \citet{nam2026worldcam,wang2026worldcompass} (see Appendix~\ref{appendix:evaluation} for details).

\paragraph{Inference Settings}
At inference time, the world model autoregressively generates the latent video in chunks of $4$ latent frames. Within each chunk, we run $50$ Euler steps under the flow-matching scheduler with time-shift $5$, and apply classifier-free guidance with scale $6$ using a precomputed null-prompt embedding for the unconditional branch. Across chunks, the key-value features of past chunks are cached and a temporally aligned $12$-frame context is selected from a $20$-frame sliding memory window for additional conditioning. All evaluations share the same inference configuration to ensure a fair comparison across method variants.

\subsection{Main Results}
\begin{table}[t]
\caption{Evaluation on \emph{visual generation quality} between different methods. Best results are in \textbf{bold}.}
\label{tab:visual}
\centering
\resizebox{\columnwidth}{!}{%
\begin{tabular}{@{}l cc cccccc@{}}
\toprule
\multirow{2}{*}{\textbf{Method}} & \multirow{2}{*}{SSIM $\uparrow$} & \multirow{2}{*}{LPIPS $\downarrow$} & \multicolumn{6}{c}{\textbf{VBench}} \\
\cmidrule(lr){4-9}
 & & & Subject $\uparrow$ & Background $\uparrow$ & Motion $\uparrow$ & Aesthetic $\uparrow$ & Imaging $\uparrow$ & Overall $\uparrow$ \\
\midrule
\rowcolor[gray]{0.95}
\textcolor{gray}{\textit{GT}} & \textcolor{gray}{\textit{1.000}} & \textcolor{gray}{\textit{0.000}} & \textcolor{gray}{\textit{0.901}} & \textcolor{gray}{\textit{0.961}} & \textcolor{gray}{\textit{0.974}} & \textcolor{gray}{\textit{0.545}} & \textcolor{gray}{\textit{0.695}} & \textcolor{gray}{\textit{0.815}} \\
\midrule
Base Model\footnotemark            & 0.458          & 0.506          & \textcolor{gray}{\textit{0.895}} & \textcolor{gray}{\textit{0.965}} & \textcolor{gray}{\textit{0.977}} & \textcolor{gray}{\textit{0.549}} & \textcolor{gray}{\textit{0.713}} & \textcolor{gray}{\textit{0.820}} \\
\quad + \textit{Random}            & 0.531          & 0.453          & 0.876          & 0.967          & 0.981          & 0.537          & 0.639          & 0.800          \\
\quad + \textit{AOL}            & \textbf{0.538} & 0.419          & \textbf{0.890} & \textbf{0.970} & \textbf{0.982} & 0.541          & 0.644          & 0.805          \\
\rowcolor[gray]{0.92}
Full Model (\textit{w/ CFT})       & 0.531          & \textbf{0.411} & \textbf{0.890} & 0.968          & 0.979          & \textbf{0.543} & \textbf{0.681} & \textbf{0.812} \\
\bottomrule
\end{tabular}%
}
\end{table}
\footnotetext{As the base model cannot effectively follow actions or produce Minecraft-style outputs, general visual metrics without reference videos (\ie, VBench scores) are not directly comparable and are listed for reference only.}

We compare our proposed method against \emph{Random}, which samples training data without active acquisition or counterfactual supervision. \tabref{tab:visual} reports the visual generation results. Visual-quality metrics are relatively close across model variants, but clear trends can still be observed. Compared with \emph{Random}, \emph{AOL} improves reconstruction fidelity and perceptual similarity, achieving higher SSIM and lower LPIPS. It also brings consistent gains across all VBench dimensions. Adding CFT further improves LPIPS and gives the best VBench Imaging and Overall scores. This suggests that counterfactual supervision complements reconstruction-based learning by improving perceptual and imaging quality. In our experiments, we observed that selecting the most difficult fork in each scene leads to significant instability during training, which confirms the need for selecting samples with balanced difficulty.

The improvement is clearer in terms of action controllability, which is the central objective of generative world modeling. Compared with \emph{Random}, \emph{AOL} improves both Combined and Fine-grained action accuracy, while reducing both rotational and translational trajectory errors. This shows that active acquisition selects samples that are more useful for learning action-conditioned transitions. With CFT, the full model achieves the best results on all controllability metrics. Relative to \emph{Random}, it improves Combined accuracy by $19.4\%$ and Fine-grained accuracy by $12.5\%$, while reducing $\mathrm{RPE}_{\mathrm{rot}}$ by $36.3\%$ and $\mathrm{RPE}_{\mathrm{trans}}$ by $20.1\%$. These gains indicate that AOL provides more effective supervision, and CFT further uses counterfactual samples to effectively strengthen the model's understanding of action transition dynamics.

\begin{table}[h]
\caption{Evaluation on \emph{action controllability} between different methods. Best results are in \textbf{bold}.}
\label{tab:control}
\centering
\resizebox{0.7\columnwidth}{!}{%
\begin{tabular}{@{}l cc cc@{}}
\toprule
\multirow{2}{*}{\textbf{Method}} & \multicolumn{2}{c}{\textbf{Action Accuracy}} & \multicolumn{2}{c}{\textbf{Camera Control}} \\
\cmidrule(lr){2-3} \cmidrule(lr){4-5}
 & Combined $\uparrow$ & Fine-grained $\uparrow$ & RPE\textsubscript{rot} $\downarrow$ & RPE\textsubscript{trans} $\downarrow$ \\
\midrule
Base Model               & 0.240 & 0.490 & 0.6906  & 0.0235  \\
\quad + \textit{Random}   & 0.356 & 0.569 & 0.6393  & 0.0199  \\
\quad + \textit{AOL}   & 0.393 & 0.610 & 0.5033  & 0.0166  \\
\rowcolor[gray]{0.92}
Full Model (\textit{w/ CFT}) & \textbf{0.425} & \textbf{0.640} & \textbf{0.4072} & \textbf{0.0159} \\
\bottomrule
\end{tabular}%
}
\end{table}
\subsection{Qualitative Results and Analysis}

\begin{figure}[t]
    \centering
    \includegraphics[width=\columnwidth]{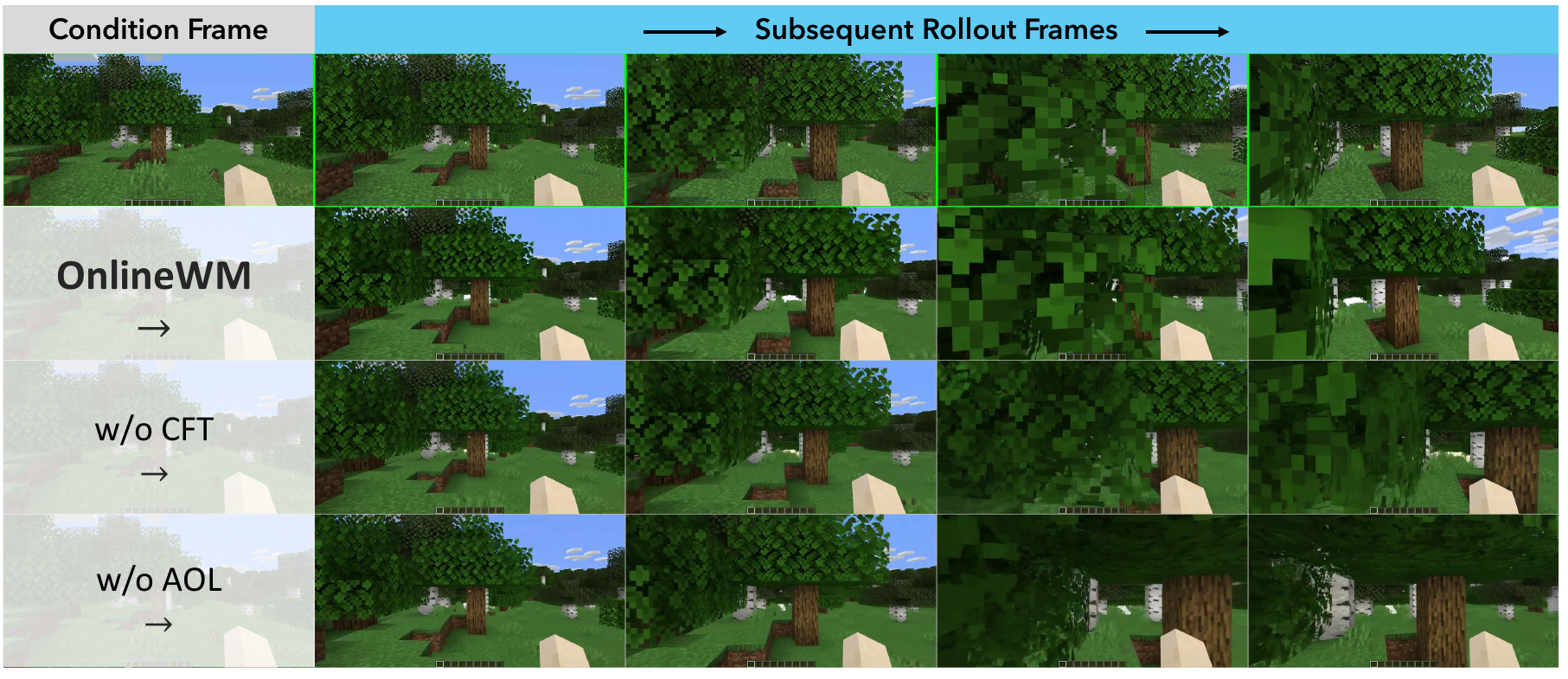}
    \caption{Visual comparison of all method variants in a challenging scenario. The first row shows the ground-truth video, and subsequent rows denote the generation results of three variants: \emph{OnlineWM}, \emph{OnlineWM w/o the CFT loss} and \emph{OnlineWM \textbf{removing the whole AOL and CFT design}}.}
    \label{fig:visualization}
\end{figure}

We further provide a qualitative evaluation of OnlineWM. \figref{fig:visualization} shows a challenging scenario where the agent moves into tree leaves, causing strong occlusion and a large visual distribution shift. OnlineWM handles this case accurately: it preserves the spatial relation between the camera, the tree trunk, and the surrounding leaves, and follows the commanded action to produce the correct collision-like transition. Removing CFT weakens this spatial reasoning, leading to less accurate alignment between the predicted motion and the scene geometry. When AOL is also removed, the model struggles with this rare and difficult transition, and the rollout collapses to a more common trajectory under the tree instead of correctly modeling the interaction with the leaves. This comparison shows that AOL helps expose the model to informative failure-prone cases, while CFT improves the learning of action-conditioned spatial dynamics. Refer to Appendix~\ref{appendix:visualization} for more qualitative examples.

\begin{figure}[!t]
    \vspace{6pt}
    \includegraphics[width=0.95\columnwidth]{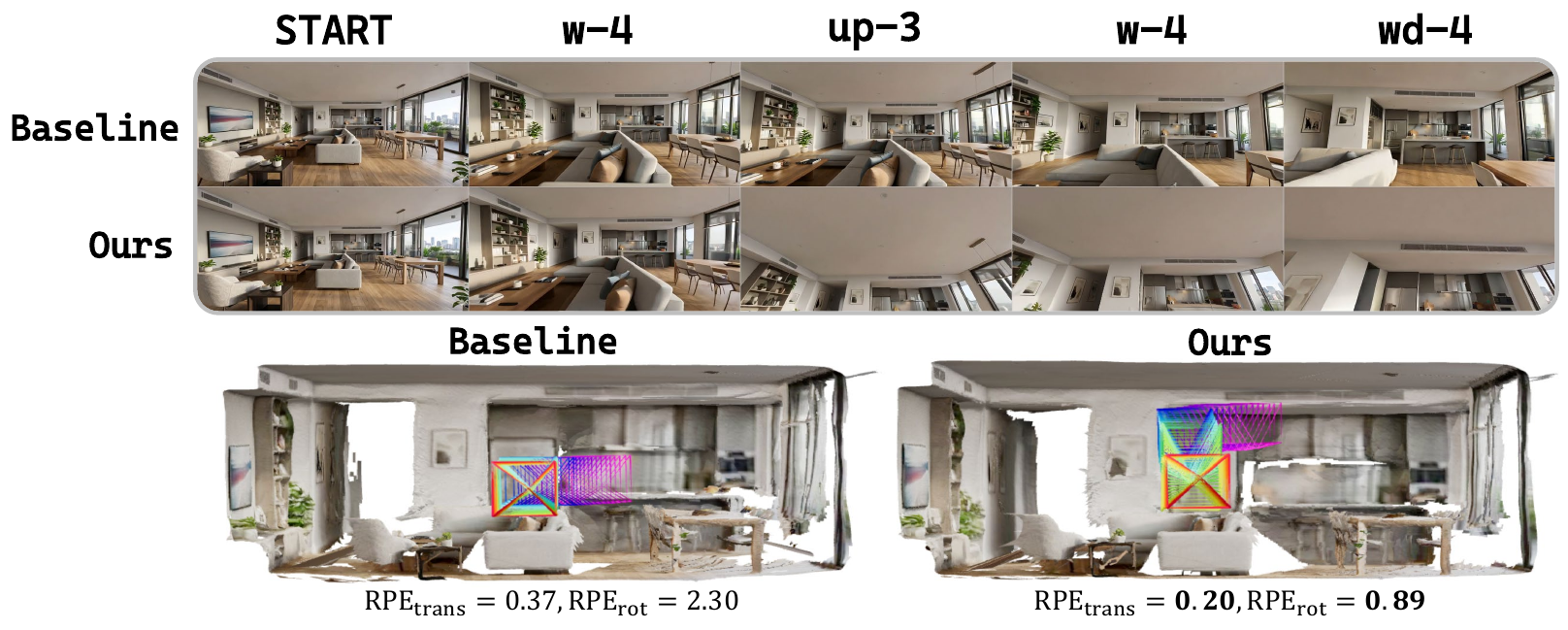}\\[12pt]
    \includegraphics[width=0.95\columnwidth]{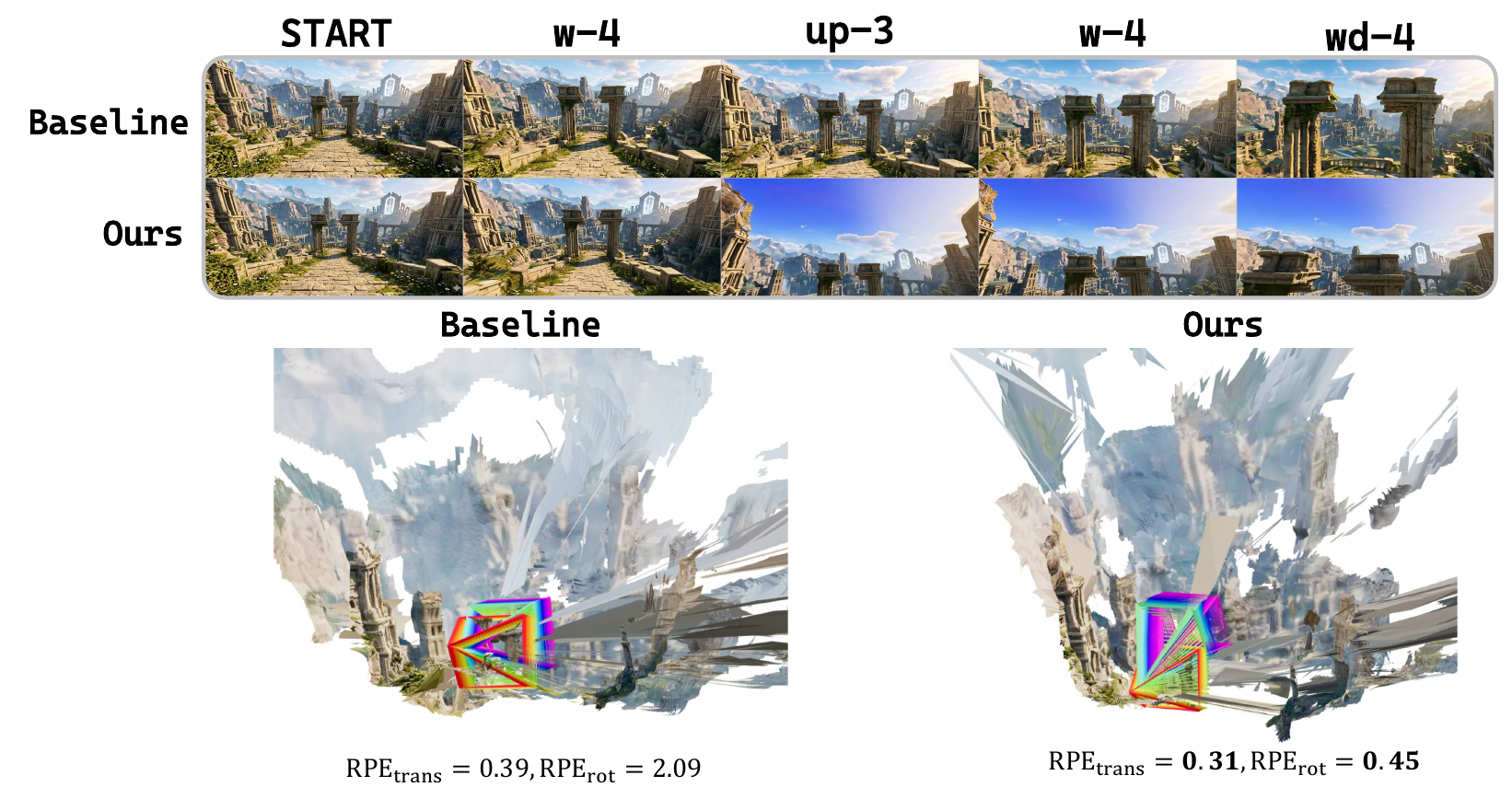}
    \caption{Visual comparison of action controllability on general domains. For each scene, the top rows show rollouts conditioned on the same action sequence (denoted by the labels at the top of each column, \eg, \texttt{w-4}, \texttt{up-3}) under the baseline and our model. The bottom row visualizes the corresponding 3D Gaussian Splatting reconstruction and the recovered camera trajectory, with relative pose errors $\text{RPE}_{\text{trans}}$ and $\text{RPE}_{\text{rot}}$ reported below.}
    \label{fig:visualization_real_world}
\end{figure}

\subsection{Generalization to General Domains}

OnlineWM learns action control through online interaction with simulators. An important question is whether the learned action-conditioned dynamics are specific to the simulator domain, or can transfer to visually different general domains. To examine this, we directly evaluate the world model trained with OnlineWM on Minecraft using general-domain images as the conditioning input, without any further training or adaptation. We compare it with the RL post-training checkpoint from WorldCompass~\citep{wang2026worldcompass}. To better stress-test action controllability, we use an action sequence with faster view changes than the default setting in WorldCompass. We report camera RPE and present the generated video alongside 3D Gaussian reconstruction results from WorldMirror~\citep{liu2025worldmirror}, annotated with the estimated camera trajectory.

As shown in \figref{fig:visualization_real_world}, the baseline model struggles to follow the conditioning actions, particularly the \texttt{up} action, which is relatively uncommon in the training data. In contrast, the model trained with OnlineWM produces view changes that are more consistent with the input actions. This improvement is reflected in both the qualitative visual comparison and the quantitative action-control metrics, where OnlineWM achieves lower translational and rotational RPE. These results demonstrate that OnlineWM does not merely improve in-domain simulator performance, but also learns action-conditioned transition knowledge that can transfer to general visual domains. It also suggests that learning effective causality-aware action control from simulators is a promising direction towards building effective general world models. Refer to Appendix~\ref{appendix:visualization} for more qualitative examples on general domains.

\FloatBarrier
\section{Conclusion}
In conclusion, OnlineWM presents a new paradigm for world model training, shifting from passive supervision to a closed-loop, active, and causality-driven refinement process. It delivers notable improvements in both action controllability and visual quality, demonstrating the fundamental benefits of online training over existing offline paradigms for world models. The generalization results further indicate that OnlineWM can serve as an effective and universal training paradigm for distilling action control from simulators into various domains. We believe this closed-loop, causality-aware perspective offers a promising path toward world models whose dynamics are grounded in reliable causal mechanisms rather than spurious visual correlations, and we hope OnlineWM can inspire future research on interactive and causally grounded learning for generative world models.

\FloatBarrier
{\small
\bibliographystyle{arxiv-numbered}
\bibliography{reference}
}

\clearpage
\appendix

\section{More Implementation Details}
\label{appendix:implementation_details}

\subsection{Model Architecture}
\label{appendix:model-architecture}

We provide additional details on the HY-World~1.5 backbone~\citep{sun2025worldplay,wu2025hunyuanvideo} used as our base world model, focusing on how discrete actions and continuous camera poses are injected into the DiT. \tabref{tab:backbone} summarizes the key architectural and VAE parameters; the rest of this section describes the conditioning mechanisms.

\begin{table}[h]
\caption{Architecture parameters of the HY-World~1.5 backbone~\citep{sun2025worldplay,wu2025hunyuanvideo}.}
\label{tab:backbone}
\centering
\small
\setlength{\tabcolsep}{8pt}
\renewcommand{\arraystretch}{1.15}
\begin{tabular}{@{}l@{\hspace{2em}}l@{}}
\toprule
\textbf{Component} & \textbf{Configuration} \\
\midrule
\rowcolor{gray!10}\multicolumn{2}{@{}l}{\emph{Diffusion transformer}} \\
Block depth (dual-stream) & $54$ \\
Hidden dim / heads / head dim & $2048$ $/$ $16$ $/$ $128$ \\
MLP expansion ratio & $4\!\times$ \\
Patch size (3D conv) & $1\!\times\!1\!\times\!1$ \\
Positional encoding & 3-axis RoPE on $(T, H, W)$, dims $(16, 56, 56)$ \\
\addlinespace[2pt]
\rowcolor{gray!10}\multicolumn{2}{@{}l}{\emph{3D causal VAE}} \\
Spatial / temporal compression factor & $16\!\times$ $/$ $4\!\times$ \\
Latent channels & $32$ \\
Latent length & $T_z = \lfloor (T-1)/4\rfloor + 1$ \\
\bottomrule
\end{tabular}
\end{table}

\paragraph{Action Injection}
Each latent frame is associated with a compound action token from the $45$-token vocabulary defined in our experimental setup. The token is embedded through the same sinusoidal-then-2-layer-MLP module that lifts the diffusion timestep, summed into the global AdaLN conditioning vector together with the timestep and pooled text embeddings, and broadcast across the visual tokens of the frame so that action information modulates every dual-stream block via AdaLN rather than entering the attention sequence as a token. The final MLP layer of the action embedder is zero-initialized, so action conditioning behaves as a no-op when adapting from a pretrained text-to-video backbone.

\paragraph{Camera Injection}
Per-frame camera poses (a $4\!\times\!4$ extrinsic and a $3\!\times\!3$ intrinsic) are injected through projective rotary positional encoding (PRoPE)~\citep{li2025cameras}, which HY-World~1.5 adopts as a parallel visual attention branch attached to each dual-stream block. Conceptually, PRoPE conditions attention on the per-frame camera projection so that geometrically corresponding rays across frames are aligned within the camera-conditioned attention space, complementing the standard 3-axis RoPE that encodes spatio-temporal position. The PRoPE branch is fused back into the visual stream through a zero-initialized linear projection and is applied only to the visual stream, while text tokens retain the standard RoPE attention.

\paragraph{Autoregressive Rollout with History}
The noisy latent chunk to be denoised is channel-wise concatenated with the latents of previously generated chunks together with a binary mask indicating observed positions, and the concatenated tensor is processed by the patch-embedding 3D convolution whose conditioning-side input channels are zero-initialized. The attention mask is bidirectional within a chunk and strictly causal across chunks, enforcing autoregressive generation while preserving full intra-chunk dependencies, and the key-value features of past chunks are cached at inference time so that only the current chunk is recomputed per rollout step. Text conditioning is processed through a separate stream within each dual-stream block and joins the visual tokens via concatenated joint attention, with separate output projections and MLPs for each modality.

\subsection{Online Data Collection and Processing Details}
\label{appendix:collection}

\paragraph{Simulator and Action Sampling}
Each parallel Minecraft world is initialized in \texttt{creative} mode with a unique random map seed and a spawn biome uniformly drawn from a fixed pool. To keep visual transitions smooth and temporally consistent with the model's chunked latent representation, the action agent samples a compound action token and holds it for a contiguous block of latent frames---roughly $0.8$--$3.2$ seconds at $20$\,FPS---before re-sampling. Longer trajectories can be composed from these action primitives. \tabref{tab:simulator_conventions} summarizes the simulator and data conventions adopted throughout our online data collection.

\begin{table}[h]
\caption{Simulator and data conventions for online data collection in OnlineWM.}
\label{tab:simulator_conventions}
\centering
\small
\setlength{\tabcolsep}{8pt}
\renewcommand{\arraystretch}{1.15}
\begin{tabular}{@{}l@{\hspace{2em}}l@{}}
\toprule
\textbf{Setting} & \textbf{Value} \\
\midrule
\rowcolor{gray!10}\multicolumn{2}{@{}l}{\emph{World and rendering}} \\
Game mode & \texttt{creative} \\
Spawn biomes (uniform draw) & \{\texttt{plains}, \texttt{forest}, \texttt{desert}, \texttt{taiga}, \texttt{savanna}\} \\
Field of view (FoV) & $70^\circ$ horizontal \\
\addlinespace[2pt]
\rowcolor{gray!10}\multicolumn{2}{@{}l}{\emph{Action and timing}} \\
View delta per simulator tick & $3^\circ$ \\
Action hold duration & $4$--$16$ latent frames ($0.8$--$3.2$\,s at $20$\,FPS) \\
World reset cadence & every $\sim\!96$ latent frames \\
\addlinespace[2pt]
\rowcolor{gray!10}\multicolumn{2}{@{}l}{\emph{Pose and conditioning}} \\
Camera extrinsic & $4\!\times\!4$ world-to-camera, first-frame normalized \\
Camera intrinsic & $3\!\times\!3$, derived from $70^\circ$ FoV at $832\!\times\!480$ \\
Caption (fixed) & \texttt{"Minecraft first-person gameplay."} \\
\bottomrule
\end{tabular}
\end{table}

\paragraph{Sample Format}
Each collected sample stores everything required to compute both the flow-matching and the CFT losses without re-running the simulator. Raw RGB frames are encoded into VAE latents on the fly and discarded, and every buffer entry consists of three parts: (i) the long mainline trajectory of $L_{\text{main}}$ chunks together with its per-latent-frame action tokens and camera poses, used for the flow-matching loss; (ii) the short positive--negative fork pair $(\mathbf{z}_+, \mathbf{z}_-)$ of $L_{\text{fork}}$ chunks each, sharing the same conditioning state $\mathbf{c}$ and used for the causality-aware loss; and (iii) the conditioning first-frame latent together with the precomputed text-encoder features for the fixed caption and vision-encoder features for the conditioning frame. Camera extrinsics are normalized to the first-frame pose so that every rollout shares a canonical origin, while intrinsics stay constant within a sample. \tabref{tab:buffer_shape} summarizes the corresponding tensor shapes.

\begin{table}[h]
\caption{Per-sample replay buffer layout in OnlineWM. Shapes use the latent spatial size $H_z\!\times\!W_z = 30\!\times\!52$ (at $832\!\times\!480$ raw resolution) and $4$ latent frames per chunk.}
\label{tab:buffer_shape}
\centering
\small
\setlength{\tabcolsep}{8pt}
\renewcommand{\arraystretch}{1.15}
\begin{tabular}{@{}l@{\hspace{1.5em}}l@{\hspace{1.5em}}l@{}}
\toprule
\textbf{Field} & \textbf{Shape} & \textbf{Description} \\
\midrule
\rowcolor{gray!10}\multicolumn{3}{@{}l}{\emph{Mainline trajectory ($L_{\text{main}}\!=\!8$ chunks)}} \\
\texttt{latents} & $[32,\, 4 L_{\text{main}},\, H_z,\, W_z]$ & post-fork mainline VAE latents \\
\texttt{action\_tokens} & $[4 L_{\text{main}}]$ & one compound token per latent frame \\
\texttt{w2c} & $[4 L_{\text{main}},\, 4,\, 4]$ & first-frame-normalized world-to-camera \\
\texttt{intrinsic} & $[4 L_{\text{main}},\, 3,\, 3]$ & derived from $70^\circ$ FoV \\
\addlinespace[2pt]
\rowcolor{gray!10}\multicolumn{3}{@{}l}{\emph{Counterfactual fork pair ($L_{\text{fork}}\!=\!1$ chunk each)}} \\
\texttt{short\_latents} & $[32,\, 4 L_{\text{fork}},\, H_z,\, W_z]$ & chosen-fork prefix, used as $\mathbf{z}_+$ in CFT \\
\texttt{neg\_latents} & $[32,\, 4 L_{\text{fork}},\, H_z,\, W_z]$ & counterfactual fork, used as $\mathbf{z}_-$ in CFT \\
Other \texttt{neg\_*} fields & shapes as in mainline & per-frame action token and pose for $\mathbf{z}_-$ \\
\addlinespace[2pt]
\rowcolor{gray!10}\multicolumn{3}{@{}l}{\emph{Shared conditioning}} \\
\texttt{image\_cond} & $[32,\, 1,\, H_z,\, W_z]$ & VAE latent of the conditioning frame \\
\texttt{prompt\_embed} & $[1{,}000,\, 3584]$ & LLM embedding of the fixed caption \\
\texttt{vision\_states} & $[729,\, 1152]$ & SigLIP embedding of the conditioning frame \\
\bottomrule
\end{tabular}
\end{table}

\subsection{Training}
\label{appendix:training}

We expand on the training procedure of OnlineWM in three parts: the hyperparameter configuration, the algorithmic skeleton of the closed-loop training pipeline, and the distributed architecture used to scale the framework across multiple nodes.

\paragraph{Hyperparameters}
\tabref{tab:hyperparameters} reports the full hyperparameter configuration used throughout our experiments, organized into optimization, distributed setup, active online learning, replay buffer, and causality-aware fine-tuning.

\begin{table}[h]
\caption{Training hyperparameters and distributed setup of OnlineWM.}
\label{tab:hyperparameters}
\centering
\small
\setlength{\tabcolsep}{8pt}
\renewcommand{\arraystretch}{1.15}
\begin{tabular}{@{}l@{\hspace{2em}}l@{}}
\toprule
\textbf{Hyperparameter} & \textbf{Value} \\
\midrule
\rowcolor{gray!10}\multicolumn{2}{@{}l}{\emph{Optimization}} \\
Optimizer & Muon~\citep{liu2025muon} (AdamW for $\le\!1$D parameters) \\
Peak learning rate & $2\!\times\!10^{-5}$ \\
LR schedule / warmup steps & \texttt{one-cycle} with cosine annealing $/$ $100$ \\
Total training steps & $3{,}000$ \\
Weight decay / max grad norm & $10^{-4}$ $/$ $1.0$ \\
Mixed precision (forward / master) & \texttt{bf16} $/$ \texttt{fp32} \\
\addlinespace[2pt]
\rowcolor{gray!10}\multicolumn{2}{@{}l}{\emph{Distributed setup}} \\
Compute & $4$ nodes $\times\, 8$ GPUs $=\, 32$ GPUs \\
Sequence-parallel groups (size per group) & $8$ ($4$) \\
Gradient accumulation steps & $2$ \\
Effective batch size & $16$ \\
Parallel simulators ($N$) & $32$ \\
\addlinespace[2pt]
\rowcolor{gray!10}\multicolumn{2}{@{}l}{\emph{Active online learning}} \\
Fork branches per world ($B$) & $8$ \\
Fork length / main rollout ($L_{\text{fork}}$ $/$ $L_{\text{main}}$) & $1$ $/$ $8$ chunks \\
Random override probability ($p_{\text{rand}}$) & $0.2$ \\
Scoring noise level ($\sigma^{\star}$) & $0.5$ \\
EMA decay (scoring \& old policy) & $0.99$ \\
AOL warmup steps & $100$ \\
\addlinespace[2pt]
\rowcolor{gray!10}\multicolumn{2}{@{}l}{\emph{Replay buffer}} \\
Capacity ($S_{\max}$) & $96$ \\
Collection interval ($N_{\text{collect}}$) & $6$ \\
Max usage per sample ($N_{\max}$) & $6$ \\
\addlinespace[2pt]
\rowcolor{gray!10}\multicolumn{2}{@{}l}{\emph{Causality-aware fine-tuning}} \\
Counterfactual mixing weight ($\lambda$) & $0.1$ \\
NFT advantage scaling ($\beta$) & $1.0$ \\
CFT loss weight ($\lambda_{\text{CFT}}$) & $0.2$ \\
Training noise schedule & \emph{shifted logit-normal} ($\mu\!=\!0,\;\sigma\!=\!1$, time-shift $3$) \\
\bottomrule
\end{tabular}
\end{table}

\paragraph{Algorithmic Skeleton}
We summarize the OnlineWM training procedure as three interleaved algorithms. Algorithm~\ref{alg:onlinewm} describes the closed-loop main pipeline, which alternates between an active online collection round and a gradient update; the collection round invokes \textbf{AOL\_Select} (Algorithm~\ref{alg:aol}) to score all $N\!\times\!B$ candidate forks under the EMA model, decompose their difficulty into scene- and action-level components, and select a chosen fork together with a counterfactual fork for each world. At each gradient step, \textbf{CFT\_Loss} (Algorithm~\ref{alg:cft}) is computed alongside the standard flow-matching loss on the long mainline trajectory and combined as $\mathcal{L}_{\text{total}} = \mathcal{L}_{\text{FM}} + \lambda_{\text{CFT}}\mathcal{L}_{\text{CFT}}$.

\begin{algorithm}[h]
\caption{OnlineWM: closed-loop active online training.}
\label{alg:onlinewm}
\begin{algorithmic}[1]
\Require Pretrained world model $\theta$ with EMA copy $\bar{\theta}$, parallel simulators $\{\mathrm{Env}_n\}_{n=1}^{N}$ at conditioning state $\mathbf{c}_n$, replay buffer $\mathcal{D}\!\leftarrow\!\emptyset$, total training steps $T$
\For{$t = 1, 2, \dots, T$}
    \If{$|\mathcal{D}| < S_{\max}$ \textbf{or} $t \bmod N_{\text{collect}} = 0$} \Comment{collection round}
        \For{$n = 1, \dots, N$}
            \State sample $B$ candidate compound action sequences and roll out for $L_{\text{fork}}$ chunks
            \State encode each fork to a latent $\mathbf{z}_{n,b}$ sharing the conditioning state $\mathbf{c}_n$
        \EndFor
        \State $\big\{\!\big(b_{+}^{(n)},\, b_{-}^{(n)}\big)\!\big\}_{n=1}^{N},\, \{s_{n,b}\} \gets$ \textbf{AOL\_Select} (Algorithm~\ref{alg:aol})
        \For{$n = 1, \dots, N$}
            \State $\mathbf{z}_{+}^{(n)} \gets \mathbf{z}_{n,\,b_{+}^{(n)}},\quad \mathbf{z}_{-}^{(n)} \gets \mathbf{z}_{n,\,b_{-}^{(n)}}$ \Comment{short positive / negative forks}
            \State continue $b_{+}^{(n)}$ in $\mathrm{Env}_n$ for $L_{\text{main}}$ chunks $\to \mathbf{z}_{\text{main}}^{(n)}$ \Comment{long mainline rollout}
            \State push $\big(\mathbf{z}_{\text{main}}^{(n)},\, \mathbf{z}_{+}^{(n)},\, \mathbf{z}_{-}^{(n)},\, \mathbf{c}_n\big)$ to $\mathcal{D}$ with priority $s_{n,\,b_{+}^{(n)}}$, evicting low-priority entries
        \EndFor
    \EndIf
    \State draw $(\mathbf{z}_{\text{main}}, \mathbf{z}_{+}, \mathbf{z}_{-}, \mathbf{c}) \sim \mathrm{Uniform}(\mathcal{D})$ and increment its usage counter \Comment{training step}
    \State $\mathcal{L}_{\mathrm{FM}} \gets$ flow-matching loss on $\mathbf{z}_{\text{main}}$ \hfill (Eq.~\ref{eq:FM})
    \State $\mathcal{L}_{\mathrm{CFT}} \gets$ \textbf{CFT\_Loss}$\big(\mathbf{z}_{+}, \mathbf{z}_{-}, \mathbf{c}\big)$ \hfill (Algorithm~\ref{alg:cft})
    \State $\theta \gets \theta - \eta\,\nabla_{\theta}\!\big(\mathcal{L}_{\mathrm{FM}} + \lambda_{\mathrm{CFT}}\,\mathcal{L}_{\mathrm{CFT}}\big)$
    \State $\bar{\theta} \gets \gamma\,\bar{\theta} + (1-\gamma)\,\theta$ \Comment{EMA update}
\EndFor
\State \textbf{return} $\theta$
\end{algorithmic}
\end{algorithm}

\begin{algorithm}[h]
\caption{\textbf{AOL\_Select}: scoring, scene/action decomposition, and rank-inverted fork selection.}
\label{alg:aol}
\begin{algorithmic}[1]
\Require fork latents $\{\mathbf{z}_{n,b}\}_{n,b}$ with shared conditioning $\{\mathbf{c}_n\}$, EMA model $\bar{\theta}$, scoring noise $\sigma^{\star}$ at timestep index $k^{\star}$, override probability $p_{\text{rand}}$
\State sample shared noise $\boldsymbol{\epsilon}\sim\mathcal{N}(\mathbf{0},\mathbf{I})$
\ForAll{$(n, b)$} \Comment{single EMA forward per fork}
    \State $\tilde{\mathbf{z}}_{n,b} \gets (1-\sigma^{\star})\,\mathbf{z}_{n,b} + \sigma^{\star}\,\boldsymbol{\epsilon}$
    \State $s_{n,b} \gets \big\lVert \mathbf{v}_{\bar{\theta}}\big(\tilde{\mathbf{z}}_{n,b},\, k^{\star},\, \mathbf{c}_{n,b}\big) - (\boldsymbol{\epsilon} - \mathbf{z}_{n,b}) \big\rVert_{2}^{2}$
\EndFor
\State $d_{n} \gets \mathrm{mean}_{b}\,s_{n,b},\quad \delta_{n,b} \gets s_{n,b} - d_{n}$ \Comment{scene / action decomposition}
\For{$n = 1, \dots, N$}
    \If{$\mathrm{rand}() < p_{\text{rand}}$}
        \State $b_{+}^{(n)} \gets$ uniform random in $\{1, \dots, B\}$
    \Else
        \State $r_{n} \gets$ percentile rank of $d_{n}$ in $\{d_{1}, \dots, d_{N}\}$ \Comment{lower $\Leftrightarrow$ easier scene}
        \State $b_{+}^{(n)} \gets$ fork in world $n$ whose action rank (by $\delta_{n,b}$) matches $1 - r_{n}$
    \EndIf
    \State $b_{-}^{(n)} \gets \arg\max_{b \neq b_{+}^{(n)}}\, \mathrm{sim}\!\big(\mathbf{z}_{n,\,b_{+}^{(n)}},\, \mathbf{z}_{n,b}\big)$ \Comment{counterfactual fork}
\EndFor
\State \textbf{return} $\big\{\!\big(b_{+}^{(n)},\, b_{-}^{(n)}\big)\!\big\}_{n=1}^{N}$ and the score table $\{s_{n,b}\}$
\end{algorithmic}
\end{algorithm}

\begin{algorithm}[h]
\caption{\textbf{CFT\_Loss}: causality-aware fine-tuning loss on a counterfactual sample pair.}
\label{alg:cft}
\begin{algorithmic}[1]
\Require positive--negative pair $(\mathbf{z}_{+}, \mathbf{z}_{-})$ sharing condition $\mathbf{c}$, current model $\mathbf{v}_{\theta}$, EMA model $\mathbf{v}_{\mathrm{old}}$, hyperparameters $\lambda$ and $\beta$
\State $h \gets \mathrm{sim}(\mathbf{z}_{+}, \mathbf{z}_{-})$, \quad $\alpha \gets h / 2$ \Comment{visual similarity (e.g., SSIM) and reconstruction weight}
\State sample $\boldsymbol{\epsilon}\sim\mathcal{N}(\mathbf{0},\mathbf{I})$ and timestep $k$ from the shifted logit-normal schedule, with noise level $\sigma_{k}$
\State $\gamma_{k} \gets \lambda\,h\,\sigma_{k}(1-\sigma_{k})$
\State $\tilde{\mathbf{z}}^{k} \gets (1-\sigma_{k})\,\mathbf{z}_{+} + \sigma_{k}\,\boldsymbol{\epsilon} + \gamma_{k}\,(\mathbf{z}_{-} - \mathbf{z}_{+})$ \Comment{counterfactual perturbation, Eq.~\ref{eq:CFT-mix}}
\State forward both networks at $(\tilde{\mathbf{z}}^{k}, k, \mathbf{c})$ to obtain $\mathbf{v}_{\theta}$ and $\mathbf{v}_{\mathrm{old}}$
\State $\mathbf{v}^{+} \gets (1-\beta)\,\mathbf{v}_{\mathrm{old}} + \beta\,\mathbf{v}_{\theta},\qquad \mathbf{v}^{-} \gets (1+\beta)\,\mathbf{v}_{\mathrm{old}} - \beta\,\mathbf{v}_{\theta}$
\State $\hat{\mathbf{z}}_{+} \gets \tilde{\mathbf{z}}^{k} - \sigma_{k}\,\mathbf{v}^{+},\qquad \hat{\mathbf{z}}_{-} \gets \tilde{\mathbf{z}}^{k} - \sigma_{k}\,\mathbf{v}^{-}$
\State $\mathcal{L}_{\mathrm{CFT}} \gets (1-\alpha)\,\big\lVert\hat{\mathbf{z}}_{+} - \mathbf{z}_{+}\big\rVert_{w}^{2} + \alpha\,\big\lVert\hat{\mathbf{z}}_{-} - \mathbf{z}_{-}\big\rVert_{w}^{2}$ \Comment{Eq.~\ref{eq:CFT-loss}}
\State \textbf{return} $\mathcal{L}_{\mathrm{CFT}}$
\end{algorithmic}
\end{algorithm}

\paragraph{Distributed Architecture}
Training runs on $4$ compute nodes with $8$ GPUs each, for a total of $32$ GPUs. The $8$B-parameter backbone is distributed via hybrid sharded data parallelism (HSDP)---sharded within each node and replicated across nodes---and combined with sequence parallelism over $8$ groups, each spanning $4$ GPUs; together with $2$ gradient accumulation steps, this yields an effective batch size of $16$ per optimizer update. The $N\!=\!32$ Minecraft simulators are evenly distributed across the four nodes, where the leader rank on each node drives its assigned simulators and broadcasts the resulting raw rollout frames to its peers, so that VAE encoding and EMA scoring run in parallel on every rank within an SP group. After each collection round, the new samples are merged into a globally synchronized replay buffer, ensuring that the uniform draw at the training step sees a consistent buffer view across all nodes.

\subsection{Inference and Evaluation}
\label{appendix:evaluation}

\paragraph{Action Accuracy via Estimated Camera Poses}
To assess action controllability, we recover discrete action tokens from the generated video and compare them against the ground truth. We first apply Depth Anything~V3~\citep{lin2025depth} to both the generated and the ground-truth videos to estimate per-frame world-to-camera poses, subsample the estimates to the latent rate, and compute the relative camera motion between consecutive latent frames. The relative translation and rotation are then independently quantized---using fixed magnitude thresholds---into a $9$-way movement label and a $5$-way view label, and combined into a single compound token $\hat{a}_n$ in the same $45$-entry space as our action vocabulary. Action controllability is measured by two complementary token-level accuracies:
\begin{equation}
    \mathrm{Acc}_{\text{combined}} \;=\; \frac{1}{N-1}\sum_{n=1}^{N-1}\mathds{1}\!\left[\hat{a}_n = a_n\right], \qquad
    \mathrm{Acc}_{\text{fine}} \;=\; \tfrac{1}{2}\big(\mathrm{Acc}_{\text{move}} + \mathrm{Acc}_{\text{view}}\big),
    \label{eq:action-acc}
\end{equation}
where $\mathrm{Acc}_{\text{combined}}$ requires both the movement and the view label to be recovered correctly (Combined in \tabref{tab:control}), while $\mathrm{Acc}_{\text{fine}}$ averages the two axis-wise accuracies (Fine-grained in \tabref{tab:control}). For each method, we sweep the translation threshold over a small candidate set and report the configuration with the highest $\mathrm{Acc}_{\text{combined}}$ following WorldCompass~\citep{wang2026worldcompass}, which compensates for the per-scene scale ambiguity inherited from the depth estimator.

\paragraph{Relative Pose Error}
For trajectory-level controllability, we report the relative pose error (RPE) between the predicted and the ground-truth camera trajectories. The predicted trajectory is first registered to the ground truth through a closed-form Sim(3) Umeyama alignment~\citep{umeyama2002least} on the camera centers, which absorbs the estimator's per-scene scale and any global frame offset. We then compare the predicted and ground-truth incremental motions of consecutive frame pairs, and report two trajectory-averaged numbers: the translational RPE, computed as the mean Euclidean norm of the residual translation between aligned and ground-truth motions, and the rotational RPE, computed as the mean geodesic angle of the residual rotation, reported in degrees.

\FloatBarrier
\section{Additional Experiments}
\label{appendix:additional_experiments}

\subsection{Additional Ablation Experiments for AOL and CFT}
\label{appendix:aol_cft_ablation}

We extend \tabref{tab:control} with two variants: \emph{Random + CFT}, which applies CFT to randomly acquired training branches, and \emph{AOL + CFT (random negative)}, which replaces the most visually similar counterfactual branch with a random alternative from the same conditioning state. All variants use the same optimizer settings, EMA decay, buffer size, inference settings, and evaluation protocol reported in \secref{sec:experiment_setup} and \tabref{tab:hyperparameters}.

\begin{table}[htbp]
\caption{Additional ablations on \emph{action controllability}. Best results are in \textbf{bold}.}
\label{tab:additional_ablation}
\centering
\resizebox{0.8\textwidth}{!}{%
\begin{tabular}{@{}l cc cc@{}}
\toprule
\multirow{2}{*}{\textbf{Method}} & \multicolumn{2}{c}{\textbf{Action Accuracy}} & \multicolumn{2}{c}{\textbf{Camera Control}} \\
\cmidrule(lr){2-3} \cmidrule(lr){4-5}
 & Combined $\uparrow$ & Fine-grained $\uparrow$ & RPE\textsubscript{rot} $\downarrow$ & RPE\textsubscript{trans} $\downarrow$ \\
\midrule
Base Model & 0.240 & 0.490 & 0.6906 & 0.0235 \\
\quad + \textit{Random} & 0.356 & 0.569 & 0.6393 & 0.0199 \\
\quad + \textit{Random + CFT} & 0.369 & 0.596 & 0.5023 & 0.0172 \\
\quad + \textit{AOL} & 0.393 & 0.610 & 0.5033 & 0.0166 \\
\quad + \textit{AOL + CFT (random negative)} & 0.410 & 0.622 & 0.4670 & 0.0163 \\
\rowcolor[gray]{0.92}
Full Model (\textit{w/ CFT}) & \textbf{0.425} & \textbf{0.640} & \textbf{0.4072} & \textbf{0.0159} \\
\bottomrule
\end{tabular}%
}
\end{table}

\paragraph{CFT without Active Acquisition}
CFT improves all four metrics under random acquisition, while yielding larger action-accuracy gains with AOL. The full model performs best across all four metrics, supporting the complementary roles of active acquisition and counterfactual supervision.

\paragraph{Counterfactual Negative Selection}
Random negatives improve all four metrics over \emph{AOL}, while visually similar negatives yield further gains. This supports using similar counterfactual outcomes to better distinguish the effects of different actions.

\FloatBarrier
\subsection{Sensitivity to the Scoring Noise Level}
\label{appendix:scoring_noise}

The scoring noise level $\sigma^{\star}$ in Eq.~\ref{eq:FM-score} directly affects the difficulty estimates used for AOL selection. We vary $\sigma^{\star}$ from $0.3$ to $0.7$ and evaluate the stability of the resulting branch rankings relative to the default $\sigma^{\star}=0.5$. \tabref{tab:scoring_noise} reports Kendall's $\tau_b$ and the percentages of rank differences within one or two positions. The ranking is stable around the default setting, with most changes occurring among branches with very similar scores, suggesting that our ranking mechanism is not overly sensitive to the particular choice of $\sigma^{\star}$.

\begin{table}[htbp]
\caption{Stability of AOL acquisition rankings across scoring noise levels, measured relative to the default $\sigma^{\star}=0.5$.}
\label{tab:scoring_noise}
\centering
\small
\setlength{\tabcolsep}{4pt}
\renewcommand{\arraystretch}{1.1}
\begin{tabularx}{\linewidth}{@{}c*{3}{>{\centering\arraybackslash}X}@{}}
\toprule
\textbf{\boldmath $\sigma^{\star}$} & \textbf{\boldmath Kendall's $\tau_b\,\uparrow$ ($p$-value)} & \textbf{\boldmath Rank difference $\leq 1\,\uparrow$} & \textbf{\boldmath Rank difference $\leq 2\,\uparrow$} \\
\midrule
0.3 & 0.8867 ($\approx 10^{-5}$) & 86.72\% & 99.22\% \\
0.4 & 0.9386 ($\approx 10^{-5}$) & 92.97\% & 100.00\% \\
\textcolor{gray}{0.5} & \textcolor{gray}{1.0000 (N.A.)} & \textcolor{gray}{100.00\%} & \textcolor{gray}{100.00\%} \\
0.6 & 0.9263 ($\approx 10^{-5}$) & 92.97\% & 99.22\% \\
0.7 & 0.8320 ($\approx 10^{-5}$) & 75.78\% & 91.41\% \\
\bottomrule
\end{tabularx}
\end{table}

\FloatBarrier
\subsection{Using the Most Difficult Fork for AOL}
\label{appendix:hardest_fork}

We also conduct a separate sampling ablation comparing random acquisition with \emph{Hardest}, which selects the \textbf{highest-scoring training fork} in each scene, $b_+^{(n)}=\arg\max_b s_{n,b}$. However, it significantly underperforms random acquisition on all six metrics and shows high training instability at early training stage (\tabref{tab:hardest_fork}), suggesting that prioritizing prediction error alone is insufficient for determining useful training data.

\begin{table}[htbp]
\caption{Comparison of random and hardest-fork acquisition.}
\label{tab:hardest_fork}
\centering
\small
\setlength{\tabcolsep}{4pt}
\renewcommand{\arraystretch}{1.1}
\begin{tabular*}{\linewidth}{@{\extracolsep{\fill}}lcccccc@{}}
\toprule
\textbf{Method} & \textbf{\boldmath SSIM $\uparrow$} & \textbf{\boldmath LPIPS $\downarrow$} & \textbf{\boldmath Combined $\uparrow$} & \textbf{\boldmath Fine-grained $\uparrow$} & \textbf{\boldmath RPE\textsubscript{rot} $\downarrow$} & \textbf{\boldmath RPE\textsubscript{trans} $\downarrow$} \\
\midrule
Random & 0.485 & 0.480 & 0.486 & 0.693 & 1.3642 & 0.2191 \\
\textit{Hardest} & 0.457 & 0.561 & 0.404 & 0.604 & 2.0324 & 0.3530 \\
\bottomrule
\end{tabular*}
\end{table}

\section{Limitations and Future Directions}
\label{appendix:limitations}
In this work, we build OnlineWM on top of Minecraft, which offers flexibility, efficiency, and rich action coverage well-suited for studying causally grounded action control. While Minecraft already supports the diverse scenarios and interactions explored in this work, modern game engines such as Unity and Unreal Engine 5 further provide photorealistic rendering, finer-grained physics, and richer motion patterns including articulated character dynamics, deformable objects, and complex environmental interactions. Bringing OnlineWM into these engines holds substantial promise for scaling up the learning of world transition dynamics and improving visual fidelity. However, it's worth noting that efficient engineering design of the online training infrastructure is also crucial to support the practicability of OnlineWM when using these advanced game engines.

Beyond modeling scene-level transition dynamics, the closed-loop and causality-driven nature of OnlineWM also points toward learning \emph{physical dynamics} through embodied interaction. Since the active online learning loop and causality-aware fine-tuning are agnostic to the specific form of action, they could in principle be extended to settings where \textbf{custom agents continuously interact with the environment in pursuit of informative experience}. For instance, a robotic manipulator might autonomously probe a simulated workspace, actively seeking states whose contact outcomes are novel yet learnable for the current world model, and constructing counterfactual rollouts—\eg, grasping versus pushing, or applying different forces from the same configuration—to anchor its predictions in genuine physical causality rather than visual co-occurrence. We see such an extension as a promising avenue for moving from passive observation of dynamics toward an interactive, embodied paradigm of physical world understanding, and leave its concrete realization to future work.

\section{More Visualization}
\begin{figure}[h]
    \centering
    \begin{subfigure}{0.90\columnwidth}
        \centering
        \includegraphics[width=\columnwidth]{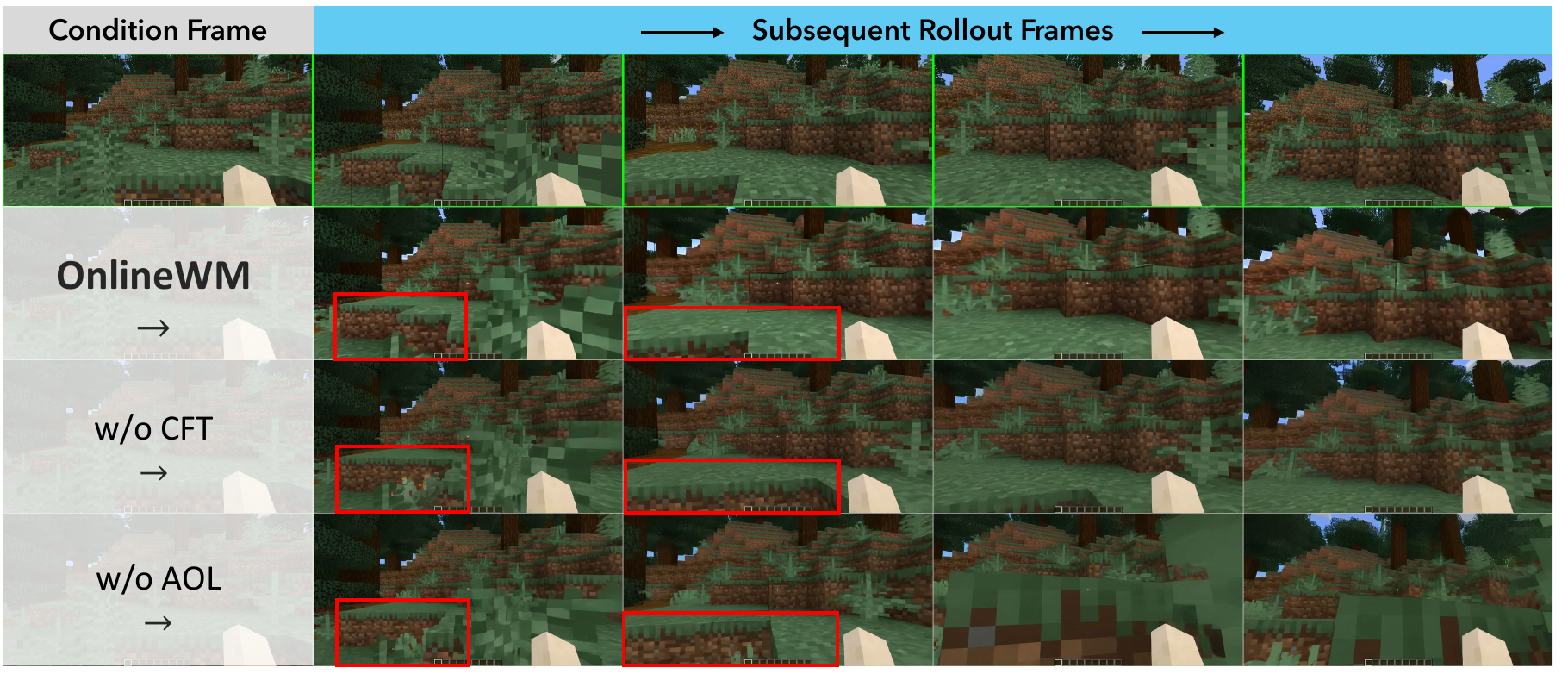}
        \caption{}
        \label{fig:visualization_appendix_1}
    \end{subfigure}
    \vspace{0.5em}
    \begin{subfigure}{0.90\columnwidth}
        \centering
        \includegraphics[width=\columnwidth]{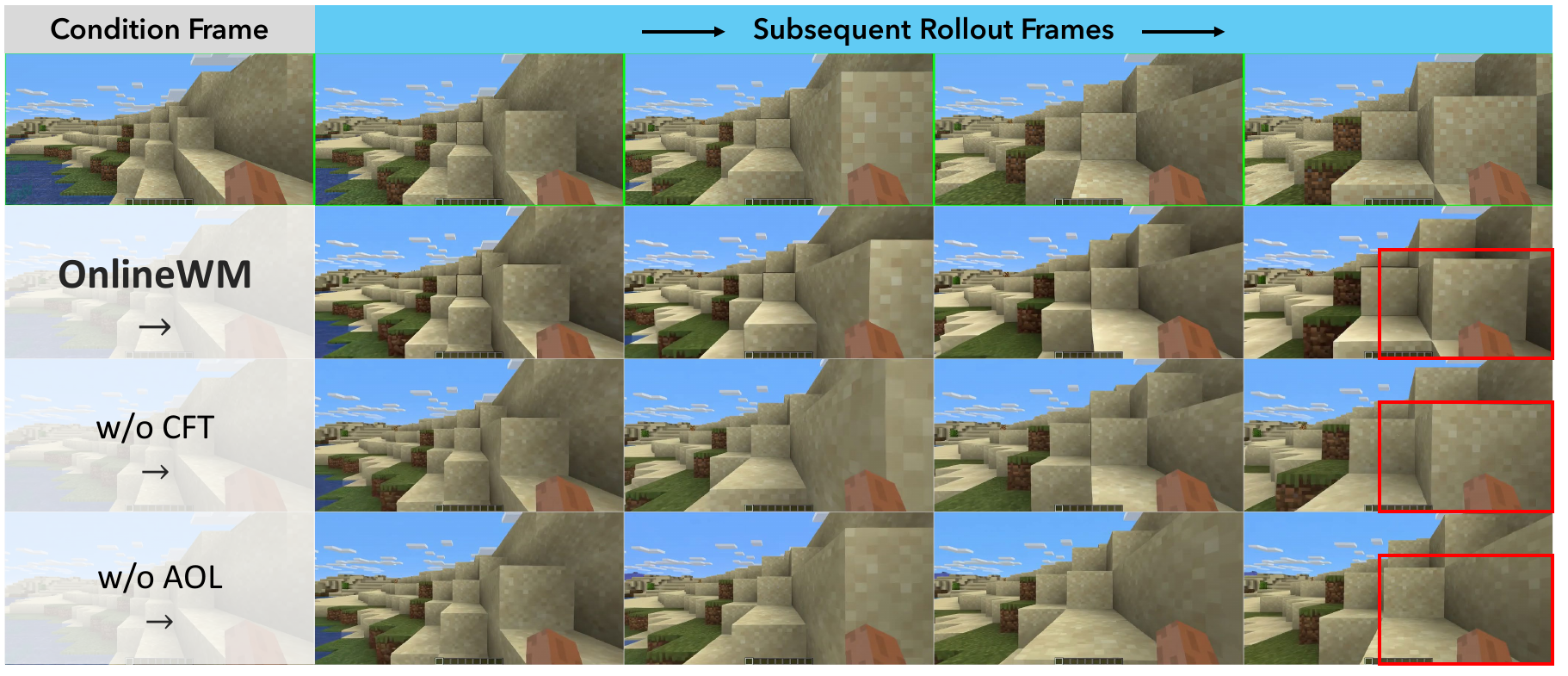}
        \caption{}
        \label{fig:visualization_appendix_2}
    \end{subfigure}
    \vspace{0.5em}
    \begin{subfigure}{0.90\columnwidth}
        \centering
        \includegraphics[width=\columnwidth]{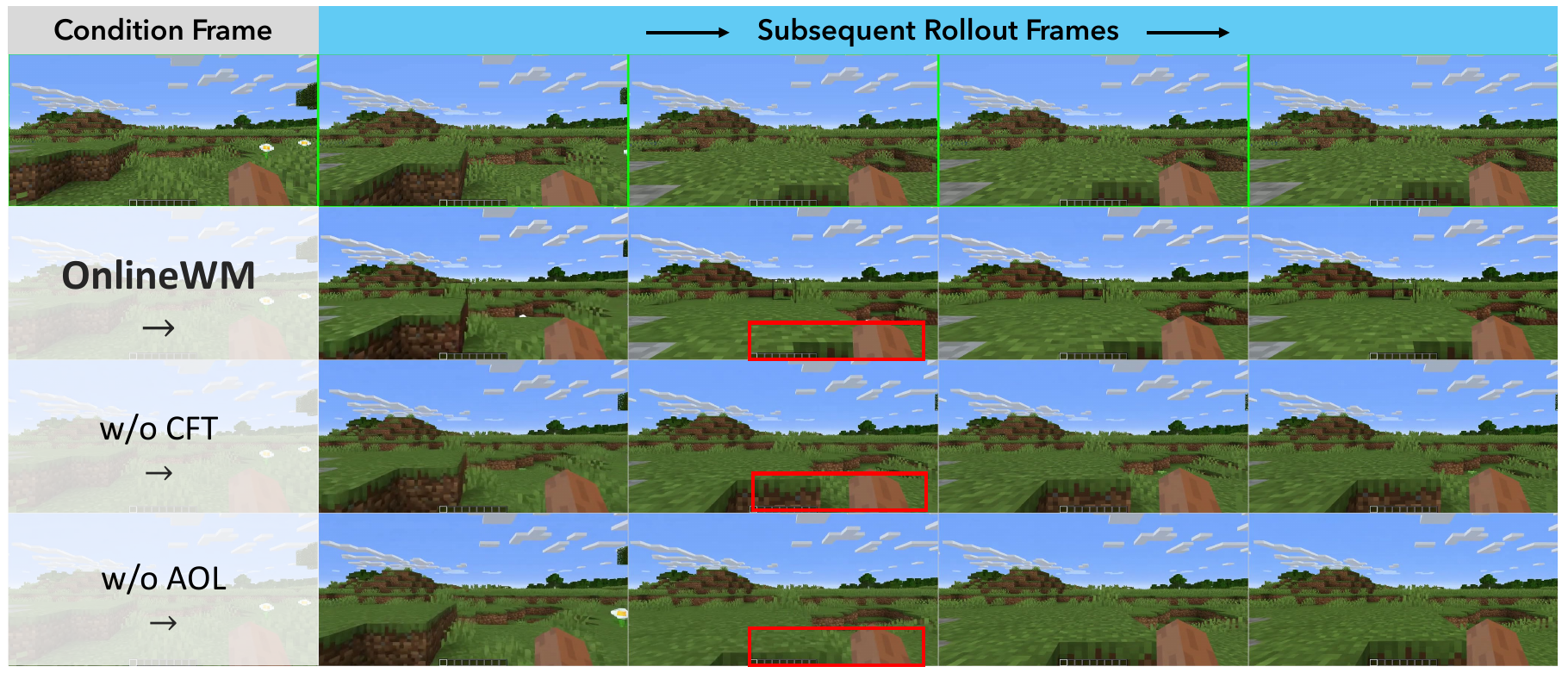}
        \caption{}
        \label{fig:visualization_appendix_3}
    \end{subfigure}
    \caption{More visual comparison of all method variants for action controllability with environmental collisions. The first row shows the ground-truth video, and subsequent rows denote the generation results of three variants: \emph{OnlineWM}, \emph{OnlineWM w/o the CFT loss}, and \emph{OnlineWM \textbf{removing the whole AOL and CFT design}}.}
    \label{fig:visualization_appendix}
\end{figure}

\begin{figure}[h]
    \includegraphics[width=0.95\columnwidth]{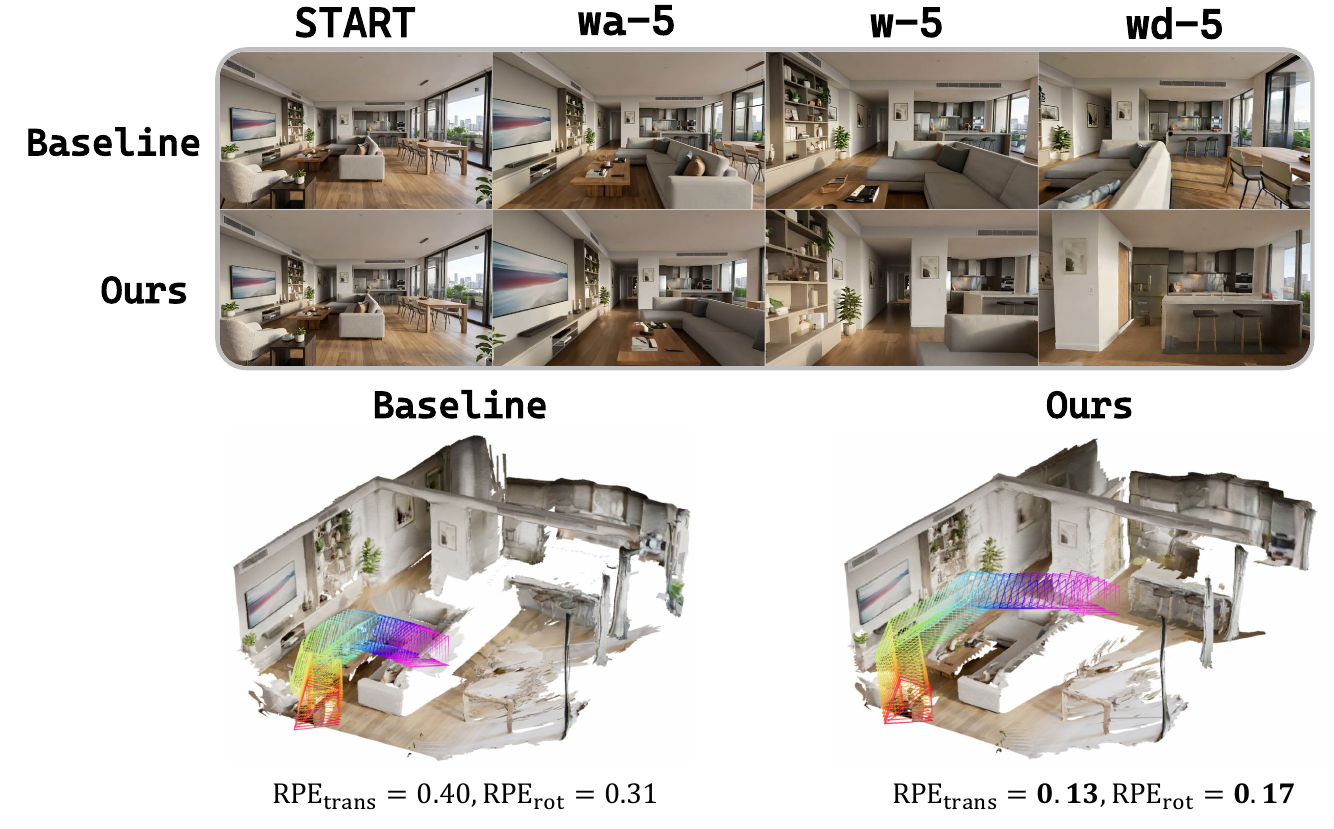}
    \caption{Visual comparison of the generalization ability of learned action control on general domains.}
    \label{fig:visualization_real_world_appendix_1}
\end{figure}

\begin{figure}[h]
    \includegraphics[width=0.95\columnwidth]{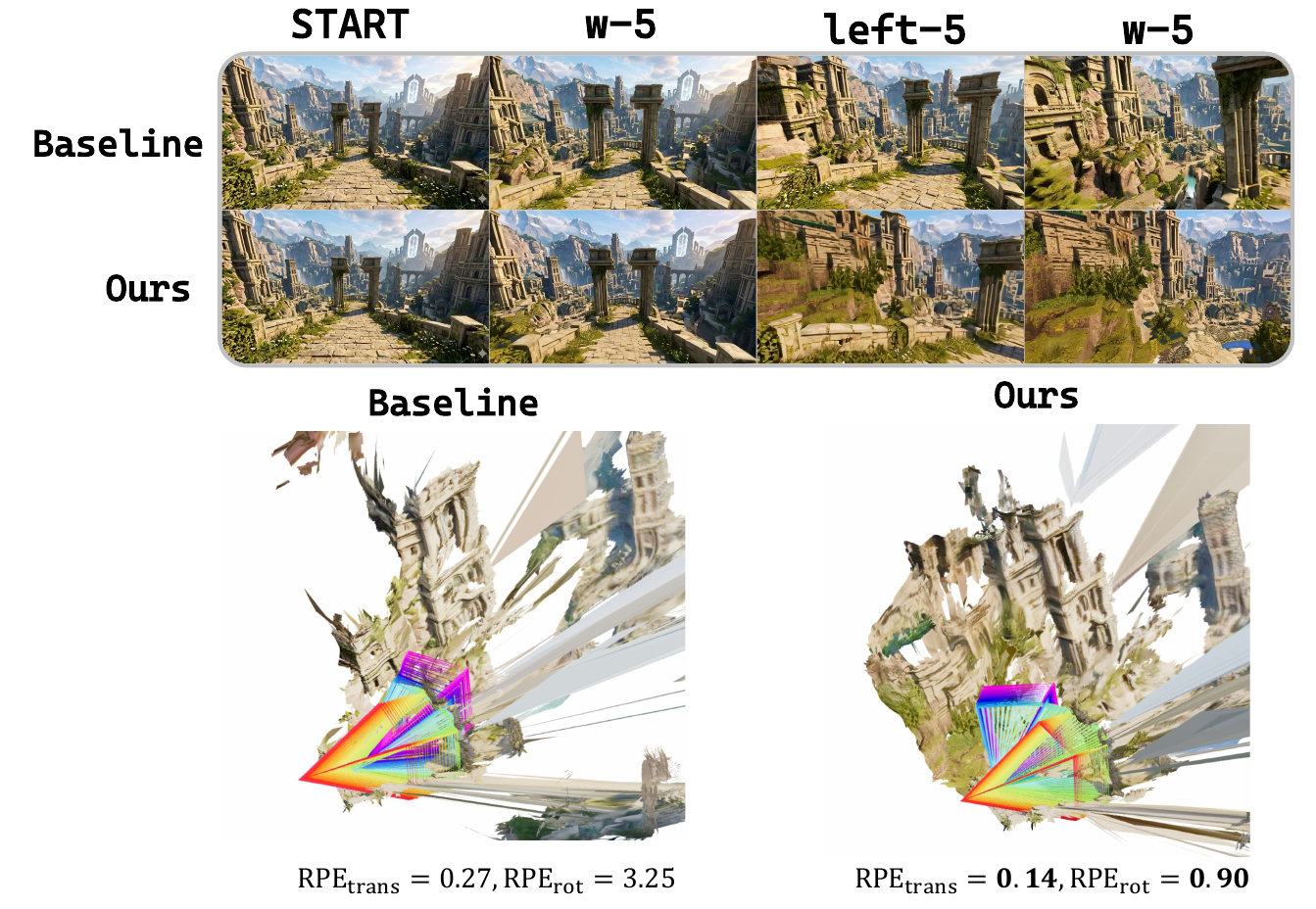}
    \caption{Visual comparison of the generalization ability of learned action control on general domains.}
    \label{fig:visualization_real_world_appendix_3}
\end{figure}

\begin{figure}[h]
    \includegraphics[width=0.95\columnwidth]{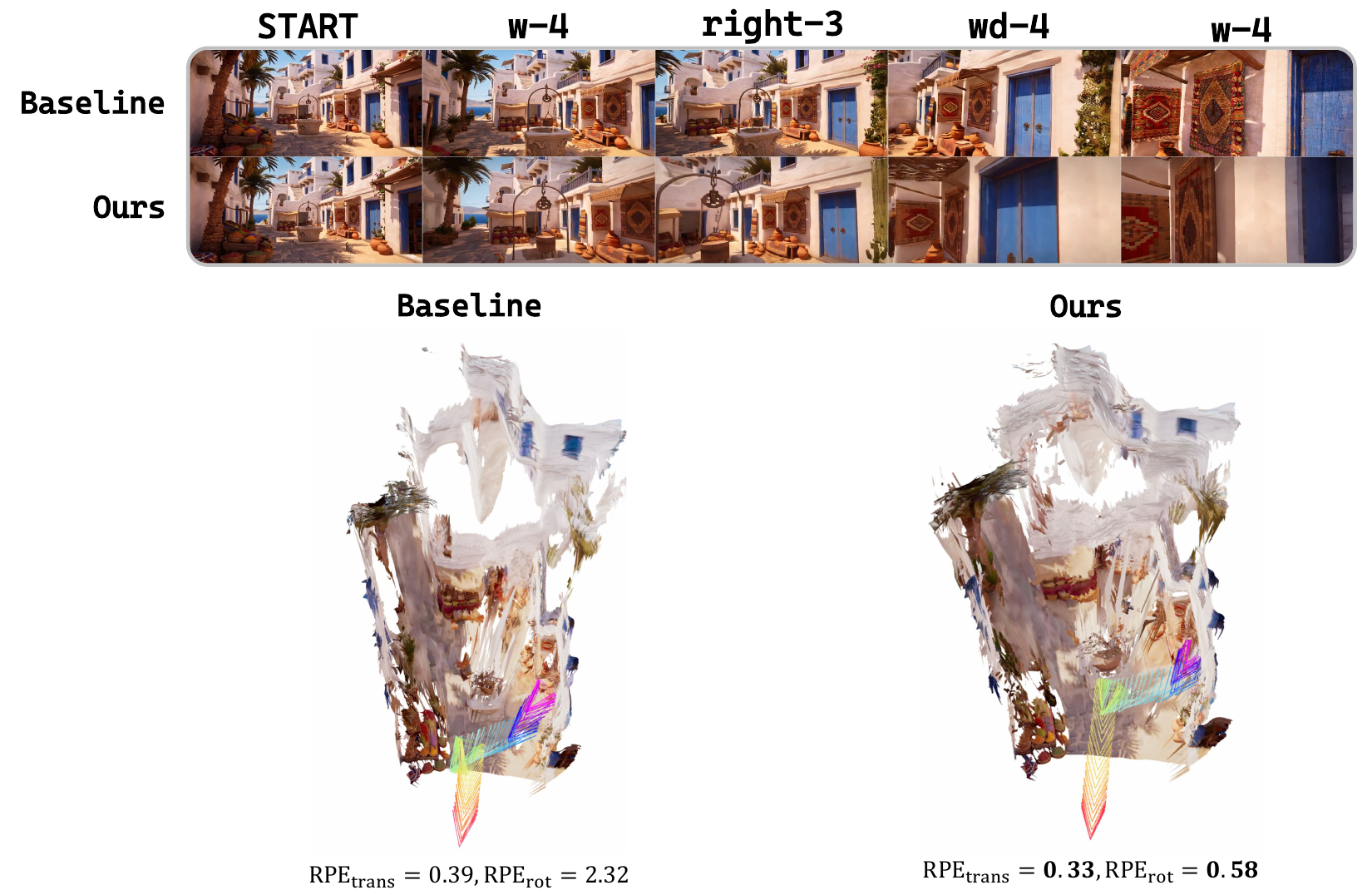}
    \caption{Visual comparison of the generalization ability of learned action control on general domains.}
    \label{fig:visualization_real_world_appendix_4}
\end{figure}

\begin{figure}[h]
    \includegraphics[width=0.95\columnwidth]{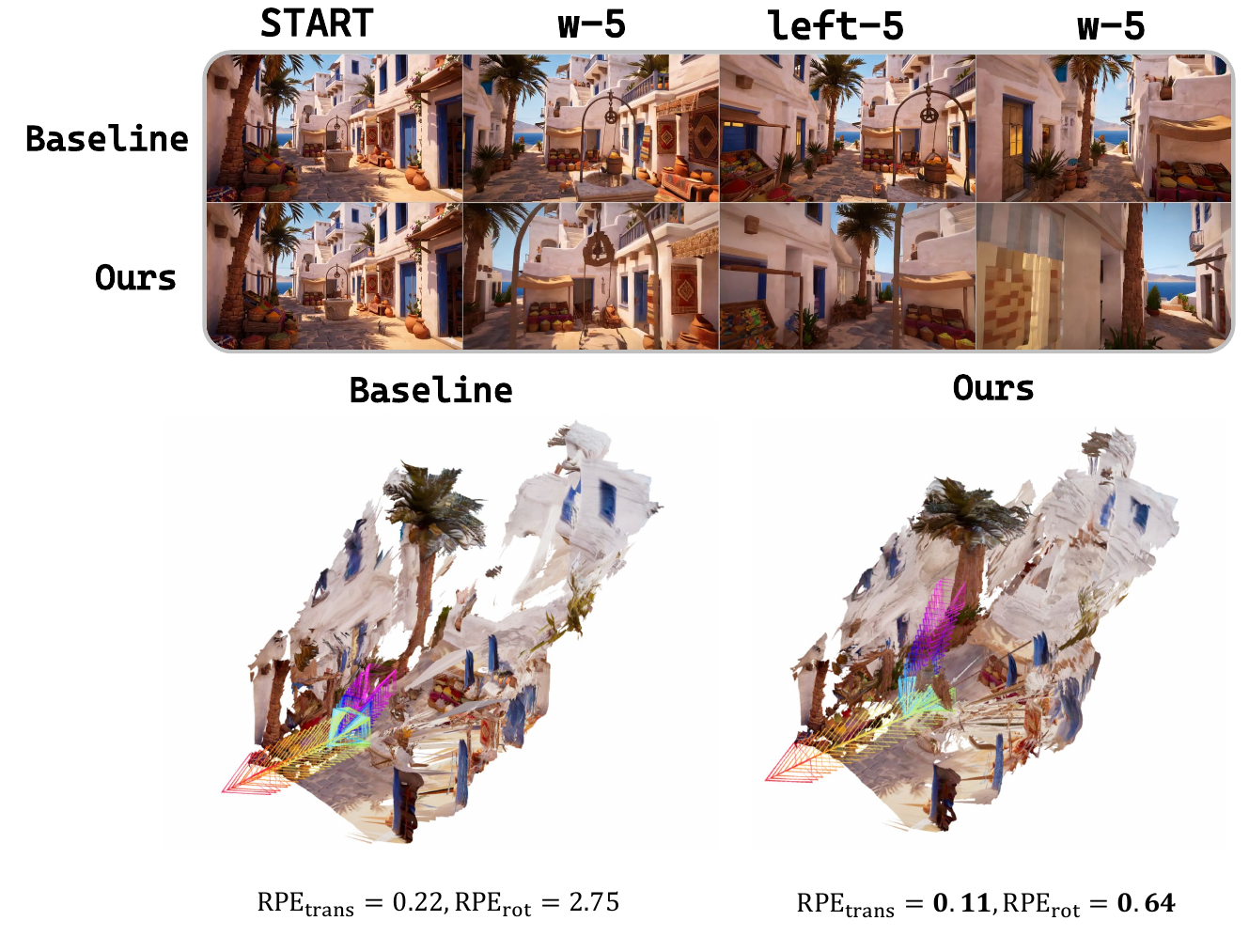}
    \caption{Visual comparison of the generalization ability of learned action control on general domains.}
    \label{fig:visualization_real_world_appendix_2}
\end{figure}

We include more qualitative visualization results in this appendix, including comparisons of action controllability for different method variants (\figref{fig:visualization_appendix}) and generalization results to general domains (\figref{fig:visualization_real_world_appendix_1}, \figrefnum{fig:visualization_real_world_appendix_2}, \figrefnum{fig:visualization_real_world_appendix_3}, \figrefnum{fig:visualization_real_world_appendix_4}).

\label{appendix:visualization}

\end{document}